\documentclass[letterpaper]{article} 
\usepackage[submission]{aaai23}  
\usepackage{times}  
\usepackage{helvet}  
\usepackage{courier}  
\usepackage[hyphens]{url}  
\usepackage{graphicx} 
\usepackage{natbib}  
\usepackage{caption} 
\usepackage{algorithm}
\usepackage{algorithmic}
\usepackage{multirow}
\usepackage{amssymb}
\usepackage{mathtools}
\usepackage{booktabs}       
\usepackage{amsfonts}       
\usepackage{xcolor}         
\usepackage{verbatimbox}
\usepackage{newfloat}
\usepackage{listings}

\DeclareCaptionStyle{ruled}{labelfont=normalfont,labelsep=colon,strut=off} 
\floatstyle{ruled}
\newfloat{listing}{tb}{lst}{}
\floatname{listing}{Listing}
\title{Compositional Scene Modeling with Global Object-Centric Representations}
\author{
    Tonglin Chen\textsuperscript{\rm },
    Bin Li\textsuperscript{\rm } \thanks{The corresponding author},
    Zhimeng Shen\textsuperscript{\rm },
    Xiangyang Xue\textsuperscript{\rm }
}
\affiliations{
    \textsuperscript{\rm } Shanghai Key Laboratory of Intelligent Information Processing, School of Computer Science, Fudan University \\
    \{tlchen18, libin, xyxue\}@fudan.edu.cn, zmshen22@m.fudan.edu.cn
}

\usepackage{bibentry}

\begin{document}

\maketitle

\begin{abstract}
The appearance of the same object may vary in different scene images due to perspectives and occlusions between objects. Humans can easily identify the same object, even if occlusions exist, by completing the occluded parts based on its canonical image in the memory. Achieving this ability is still a challenge for machine learning, especially under the unsupervised learning setting. Inspired by such an ability of humans, this paper proposes a compositional scene modeling method to infer global representations of canonical images of objects without any supervision. The representation of each object is divided into an intrinsic part, which characterizes globally invariant information (i.e. canonical representation of an object), and an extrinsic part, which characterizes scene-dependent information (e.g., position and size). To infer the intrinsic representation of each object, we employ a patch-matching strategy to align the representation of a potentially occluded object with the canonical representations of objects, and sample the most probable canonical representation based on the category of object determined by amortized variational inference. Extensive experiments are conducted on four object-centric learning benchmarks, and experimental results demonstrate that the proposed method not only outperforms state-of-the-arts in terms of segmentation and reconstruction, but also achieves good global object identification performance.
\end{abstract}

\section{Introduction}\label{sec:intro}
Object occlusion is a common phenomenon in the real world. The human visual system perceives partially occluded objects differently from complete objects \citep{hegde2008preferential}. Numerous studies in cognitive psychology and neurocomputing have shown that when recognizing occluded objects, the human visual system automatically completes the occluded parts \cite{rauschenberger2001masking,kellman1998common,behrmann1998object}. The underlying mechanism may be that the human visual system is able to imagine the occluded parts by comparing the local features of the occluded object with the previously seen complete object \citep{lerner2002object}. Humans can recognize the same objects with different occlusions, while current computer vision methods cannot achieve the similar results without supervision. Recognition of the same objects in multi-object scenes is still a huge challenge under the unsupervised setting.

The object-centric representation learning (ORL) methods which are developed based on deep generative models like Generative Adversarial Network (GAN) \citep{goodfellow2014generative} and Variational Autoencoder (VAE) \citep{kingma2013auto} can be used to learn compositional scene representations in an unsupervised manner, and have gained increasing research interests in recent years. Most methods, such as SPACE \citep{lin2019space}, GMIOO \citep{yuan2019generative}, MONet \citep{burgess2019monet}, IODINE \citep{greff2019multi}, GENESIS \citep{engelcke2019genesis}, and Slot Attention \citep{locatello2020object}, can learn representations of objects from multi-object scenes, but the learned representations of the same object with different occlusions may be quite different. Therefore, these methods are unable to learn global representations that can be used to identify potentially occluded objects in multi-objects visual scenes.

\begin{figure}
\centering
\includegraphics[width=0.75\columnwidth]{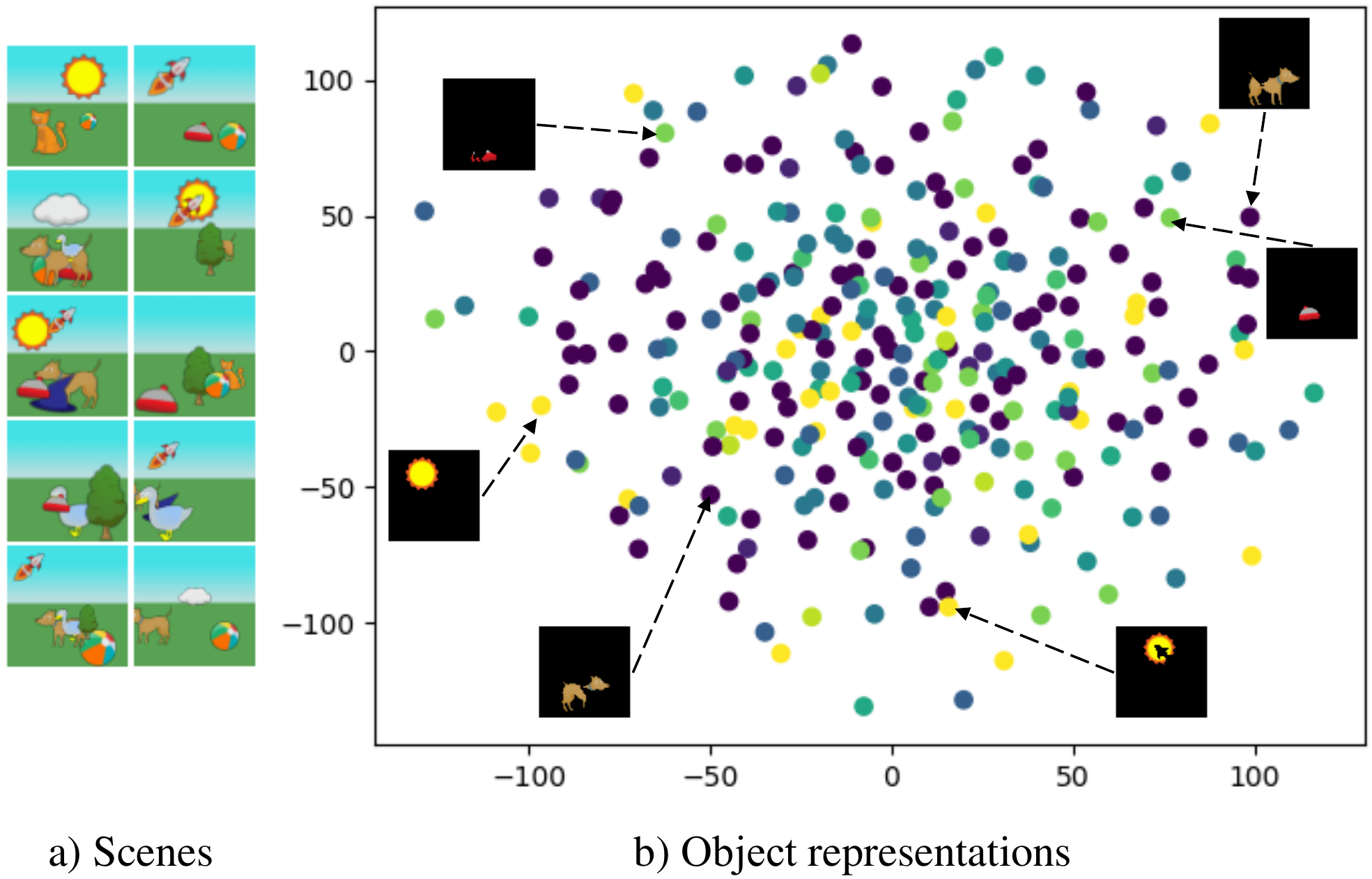}
\caption{Visualization of the object representations. }
\label{fig:different_latent}
\end{figure}

As demonstrated in Figure \ref{fig:different_latent}, recognizing the same objects with different occlusions is a huge challenge for deep generative models, since the representations of the same object with different occlusions may vary greatly. In Figure \ref{fig:different_latent}(b), the points with the same color indicate instances of the same object. The object representations is obtained by training a GENESIS-V2 \cite{engelcke2021genesis} on a multi-object scene dataset, some examples of which are shown in Figure \ref{fig:different_latent}(a). The result shown in Figure \ref{fig:different_latent}(b) is obtained by embedding the object representation to a 2D space via t-SNE \cite{van2008visualizing}. One can see that the points representing different objects are mixed together in the 2D space, which implies that the latent variable representations of different objects may be close to each other, while the latent variable representations of the same object with different occlusions may be far apart. Therefore, it is challenging to recognize the same object based on the representations extracted from the images.

To overcome this hurdle, we propose an object-centric representation learning method, called \textbf{G}lobal \textbf{O}bject-\textbf{C}entric \textbf{L}earning (GOCL), to learn global representations of canonical images of objects without any supervision. More specifically, the representations of objects are divided into two parts that characterize the intrinsic (e.g. appearance and shape) and extrinsic (e.g. position and size) attributes of objects. The representation of intrinsic attributes of the same object (i.e. canonical representation of a complete object) is globally invariant in all scene images, while the representation of extrinsic attributes is scene-dependent. Object identification can be achieved by aligning the intrinsic representation of a possibly occluded object in the current scene to the canonical representation of complete objects via a patch-matching strategy. The canonical representations of complete objects are learned by reconstruction loss in the framework of amortized variational inference. 
 
To the best of our knowledge, no existing object-centric learning method can recognize the same objects well from multi-object scenes when occlusions exist, which demonstrates the contribution of the proposed method. The proposed GOCL is compared with four representative object-centric learning methods, i.e., GENESIS-V2  \cite{engelcke2021genesis}, SPACE \cite{lin2019space}, MarioNette \cite{smirnov2021marionette} and DTI-Sprites \cite{monnier2021unsupervised} in terms of image segmentation, image reconstruction, as well as object identification. In addition, the quality of the canonical representations of objects learned by GOCL is also evaluated and compared through visualizing the generated canonical objects. Experimental results on four synthetic datasets have validated the effectiveness of the proposed GOCL.

\section{Related Works}
In recent years, several object-centric representation learning methods have been proposed to learn compositional scene representations in an unsupervised manner. The main idea of object-centric representation learning is that a visual scene image can be modeled as the composition of multiple objects, and the representation of the entire visual scene can be obtained by learning the representation of each object appeared in the visual scene. Existing object-centric representation learning methods can be divided into two categories according to the modeling of visual scenes  \citep{yuan2022compositional}. Methods in one category, e.g., Tagger \cite{greff2016tagger}, RC \cite{greff2015binding}, N-EM \cite{greff2017neural}, LDP \cite{yuan2019spatial}, GMIOO \cite{yuan2019generative}, IODINE \cite{greff2019multi}, MONet \cite{burgess2019monet}, ADI \cite{yuan2021knowledge}, GENESIS \cite{engelcke2019genesis} and OCLOC \cite{yuan2022unsupervised} model visual scenes with spatial mixture models. Methods in the other category, e.g., AIR \cite{eslami2016attend}, SPAIR \cite{crawford2019spatially}, SPACE \cite{lin2019space}, SQAIR \cite{kosiorek2018sequential}, GNM \cite{zhong2021neural}, and G-SWM \cite{lin2020improving}, use weighted summations to composite the layers of objects and background.

Methods like Slot Attention \cite{locatello2020object}, EfficientMORL \citep{emami2021efficient}, and GENESIS-V2 \cite{engelcke2021genesis} learn object representations by clustering local features of the scene image with attention mechanisms. Slot Attention randomly initializes the representations and computes and uses a Gated Recurrent Unit (GRU) \cite{cho2014learning} to iteratvely refine the representations based on the similarities between the representations and the local features of scene image. EfficientMORL \citep{emami2021efficient} models latent variables with a hierarchical probabilistic model based on Slot Attention, and gradually reduces the iterations of inference to improve efficiency. GENESIS-V2 \citep{engelcke2021genesis} learns soft cluster assignments for each pixel in a way inspired by Instance Colouring Stick-Breaking Process, and infers latent variables without iterative refinement. The proposed method can not only learn the representations of objects in the current scene using the attention mechanism like the above, but also learn globally invariant representations of objects, which can hardly be achieved by these methods.

Methods such as DTI-Sprites \cite{monnier2021unsupervised}, PCDNet \cite{villar2021unsupervised}, GSGN \cite{Deng2021Generative}, and MarioNette \cite{smirnov2021marionette} are  able to learn prototypes from visual scenes.  DTI-Sprites and PCDNet first predict the transformation parameters of each object in the scene with neural networks, and then transform the images of learnable prototpyes to reconstruct the scene image. GSGN and MarioNette learn a dictionary of prototype embeddings based on recurring patterns in the scene. By learning a decoder, the corresponding patterns can be generated from the prototype vectors to reconstruct the scene. Although these methods can also learn globally invariant prototypes, they are quite different from the proposed method. For example, DTI-Sprites and PCDNet can only learn prototype images (e.g. templates), and cannot directly obtain the representations of prototypes. Although MarioNette can learn the representations of prototypes, each image decoded from the representations may be a local patch of a complete object, or contain multiple objects, while each image decoded by the globally invariant representation in the proposed method is a complete object. The proposed method can identify the same objects with different occlusions in different scene images, which is difficult for these methods.


\section{Methodology}
Let $\boldsymbol{x}$ denote the image of a visual scene, $K$ denote the maximum number of objects that may appear in each image, $C$ denote the number of categories of objects. The attributes of objects are divided into three parts, i.e. one for intrinsic attributes such as appearance and shape, one for extrinsic attributes such as scale and translation, and the other for the category of object. The first two parts are represented by continuous variables, and the last part is represented by a discrete variable. The representation that characterizes intrinsic attributes is determined by the category of the object, and is chosen from one of the $C$ canonical representations that are shared across all the scene images.

\subsection{Generative Model}
Each image $\boldsymbol{x}$ is assumed to contain at most $K$ objects on  the background. The latent variables indicating extrinsic attributes of objects are sampled from normal distributions. According to the categorical latent variable sampled from the prior categorical distribution, one of the $C$ global intrinsic representations is selected as the representation of the intrinsic attributes of a possibly occluded object. The details of the generative model are described below.
\begin{align*}
    \boldsymbol{z}^{\text{bck}} & \sim \text{Normal}(\boldsymbol{0},\boldsymbol{I}), \\
    \boldsymbol{z}_{k}^{\text{ext}} & \sim \text{Normal}(\boldsymbol{\mu}^{\text{ext}},\boldsymbol{\sigma}^{\text{ext}}), \\
    y_{k} &\sim \text{Categorical}(\pi_1,\dots, \pi_C), \\
    \boldsymbol{z}_{k}^{\text{int}} &= \boldsymbol{e}_{y_{k}}^{\text{can}}\\
     \boldsymbol{a}_{n,0}, \boldsymbol{m}_{n,0} &=  g_{\text{dec}}^{\text{bck}}(\boldsymbol{z}^{\text{bck}})\\
     \boldsymbol{a}_{n,k}, \boldsymbol{m}_{n,k} &=f_{\text{STN}}^{-1}(g_{\text{dec}}^{\text{obj}}(\boldsymbol{z}_{k}^{\text{int}}), \boldsymbol{z}_{k}^{\text{ext}})\\
    \boldsymbol{\hat{m}}_{n,0},\dots,  \boldsymbol{\hat{m}}_{n,K} &= \text{softmax}(\boldsymbol{m}_{n,0},\dots, \boldsymbol{m}_{n,K}),\\
    \boldsymbol{x}_{n} &\sim \sum\nolimits_{k=0}^{K}\boldsymbol{\hat{m}}_{n,k} \text{Normal}\big(\boldsymbol{a}_{n,k}, \sigma^{2}\boldsymbol{I}\big)
\end{align*}
In the above expressions, some of the ranges of indexes $n~(1\leq n \leq N)$ and $ k~ (1 \leq k \leq K$ for $\boldsymbol{a},\boldsymbol{m},\boldsymbol{\hat{m}}, \boldsymbol{z}^{\text{ext}}, \boldsymbol{y}, \boldsymbol{z}^{\text{int}})$ are omitted for notational simplicity. $N$ is the total number of pixels in image $\boldsymbol{x}$. $\boldsymbol{z}^{\text{bck}}$ sampled from a normal distribution denotes the latent variable representation of background and can be converted to the appearance $\boldsymbol{a}_{0}$ and shape $\boldsymbol{m}_{0}$ of background via a neural networks $g_{\text{dec}}^{\text{bck}}$.  $\boldsymbol{z}_{k}^{\text{ext}} \in \mathbb{R}^{4}$ is the latent variable indicating the extrinsic attribute of the $k$th object, and is sampled from a normal distribution with hyperparameters $\boldsymbol{\mu}^{\text{ext}}$ and $\boldsymbol{\sigma}^{\text{ext}}$. $y_k \in \{1, 2, \dots, C\}$ is the latent variable indicating the category of the $k$th object, and is sampled from a categorical distribution with learnable parameters $\{\pi_1, \dots, \pi_C\}$. $\boldsymbol{z}_{k}^{\text{int}} \in \mathbb{R}^{D}$ is the embedding that represents the intrinsic attribute of the $k$th object, and is selected from $C$ learnable canonical representations $\{\boldsymbol{e}_{1}^{\text{can}}, \dots,\boldsymbol{e}_{C}^{\text{can}}\}$ according to the category $y_k$ of the object. $\boldsymbol{a}_{n,k}$ and $\boldsymbol{m}_{n,k} (1\leq k\leq K)$ are the respective expectations of complete appearance and unnormalized mask of the $k$th object at the $n$th pixel, and can be obtained by transforming $\boldsymbol{z}_{k}^{\text{int}}$ and $\boldsymbol{z}_{k}^{\text{ext}}$ with the combination of a decoder network $g_{\text{dec}}^{\text{obj}}$ and a Spatial Transform Network (STN) \cite{jaderberg2015spatial} $f_{\text{STN}}^{-1}$. $\boldsymbol{\hat{m}}_{n,k}$ is the perceived shape of the $k$th object at the $n$th pixel, and is computed by normalizing $\{\boldsymbol{m}_{1}, \boldsymbol{m}_{2}, \dots, \boldsymbol{m}_K\}$ with the softmax function. $\boldsymbol{x}$ is the generated image, and is sampled from a spatial mixture model parameterized by $\hat{\boldsymbol{m}}$, $\boldsymbol{a}$, and a hyperparameter $\sigma$.

The joint probability distribution of the image $\boldsymbol{x}$ and all the latent variables $\boldsymbol{\Omega} = \{ \boldsymbol{z}^{\text{ext}},\boldsymbol{y},\boldsymbol{z}^{\text{bck}}\}$ is
\begin{equation}
    p(\boldsymbol{x},\!\boldsymbol{\Omega}; \boldsymbol{e}_{1:C}^{\text{can}}) \!=\!p(\boldsymbol{z}^{\text{bck}})\!\prod_{k=1}^{K}\!p(y_{k})p(\boldsymbol{z}_{k}^{\text{ext}}) p(\boldsymbol{x}|\boldsymbol{\Omega}; \boldsymbol{e}_{1:C}^{\text{can}})
 \end{equation}

\subsection{Variational Inference}
All the latent variables $\boldsymbol{\Omega} = \{ \boldsymbol{z}^{\text{ext}},\boldsymbol{y},\boldsymbol{z}^{\text{bck}}\}$ are inferred by amortized variational inference. The intractable posterior distribution is approximated with the variational distribution whose parameters are estimated by transforming the observed image $\boldsymbol{x}$ with neural networks. In the variational distribution, latent variables $\boldsymbol{z}^{\text{ext}}$ and $\boldsymbol{y}$ of each object are assumed to be independent of each other, and the factorization of the variational distribution is
\begin{equation}
    \label{equ:infer}
    \!q(\boldsymbol{\Omega} |\boldsymbol{x};\!\boldsymbol{e}_{1:C}^{\text{can}}) \!=\!q(\boldsymbol{z}^{\text{bck}}|\boldsymbol{x}) \!\!\prod_{k=1}^{K}\!\! q(\boldsymbol{z}_{k}^{\text{ext}}|\boldsymbol{x})q(y_{k}|\boldsymbol{z}_{k}^{\text{ext}},\!\boldsymbol{x};\!\boldsymbol{e}_{1:C}^{\text{can}})
\end{equation}
The specific forms of the distributions on the right-hand side of Eq. \eqref{equ:infer} are
\begin{align*}
    q(\boldsymbol{z}^{\text{bck}}|\boldsymbol{x}) &= \text{Normal}\Big(\hat{\boldsymbol{\mu}}^{\text{bck}},\text{diag}\big((\hat{\boldsymbol{\sigma}}^{\text{bck}})^2\big)\Big)\\
    q(\boldsymbol{z}_{k}^{\text{ext}}|\boldsymbol{x}) &= \text{Normal}\Big(\hat{\boldsymbol{\mu}}_{k}^{\text{ext}},\text{diag}\big((\hat{\boldsymbol{\sigma}}_{k}^{\text{ext}})^2\big)\Big)\\
    q(y_{k}|\boldsymbol{z}_{k}^{\text{ext}},\boldsymbol{x};\boldsymbol{e}_{1:C}^{\text{can}}) &= \text{Categorical}(\hat{\pi}_{k,1},\dots,\hat{\pi}_{k,C})
\end{align*}
In the above expressions, parameters  $\hat{\boldsymbol{\mu}}^{\text{bck}}$ and $\hat{\boldsymbol{\sigma}}^{\text{bck}}$ are estimated by a background encoder $f_{\text{enc}}^{\text{bck}}$ with scene image $\boldsymbol{x}$ and the attention mask $\boldsymbol{m}_{0}^{\text{enc}}$ of the background as inputs. Parameters $\hat{\boldsymbol{\mu}}_{k}^{\text{ext}}$ and $\hat{\boldsymbol{\sigma}}_{k}^{\text{ext}}$ are estimated via an encoder network $f_{\text{enc}}^{\text{ext}}$ whose inputs are the concatenation of the scene image $\boldsymbol{x}$ and the attention mask $\boldsymbol{m}_{k}^{\text{enc}} (1 \leq k \leq K)$ of the $k$th foreground object. The attention masks $\boldsymbol{m}_{0:K}^{\text{enc}}$ are calculated using  $f_{\text{enc}}^{\text{mask}}$ which is modified function based on the IC-SBP proposed in GENESIS-V2 \citep{engelcke2021genesis}. In IC-SBP, the cluster centers of objects and background are randomly sampled from a uniform distribution, resulting in that the attention mask of the background cannot be distinguished from those of objects. In $f_{\text{enc}}^{\text{mask}}$, the cluster center of the background is first determined by calculating the sum of the distances between the pixel embedding belonging the background and all the pixel embeddings in the scene.  When the cluster center of the background is determined, the attention masks of the background and foreground objects can be calculated in the same way as IC-SBP. The details of $f_{\text{enc}}^{\text{mask}}$ will be described in Supplementary Materials. The computation of $\hat{\boldsymbol{\mu}}^{\text{bck}}$, $\hat{\boldsymbol{\sigma}}^{\text{bck}}$, $\hat{\boldsymbol{\mu}}_{k}^{\text{ext}}$ and $\hat{\boldsymbol{\sigma}}_{k}^{\text{ext}}$ is shown below.
\begin{align}
    \label{equ:infermask}
    \boldsymbol{m}_{0:K}^{\text{enc}} & = f_{\text{enc}}^{\text{mask}}(\boldsymbol{x}) \\
    \label{equ:bck}
    \hat{\boldsymbol{\mu}}^{\text{bck}},\hat{\boldsymbol{\sigma}}^{\text{bck}} &= f_{\text{enc}}^{\text{bck}}([\boldsymbol{x},\boldsymbol{m}_{0}^{\text{enc}}]) \\
    \label{equ:ext}
    \hat{\boldsymbol{\mu}}_{k}^{\text{ext}}, \hat{\boldsymbol{\sigma}}_{k}^{\text{ext}} & = f_{\text{enc}}^{\text{ext}}([\boldsymbol{x},\boldsymbol{m}_{k}^{\text{enc}}])
\end{align}
where $\boldsymbol{m}_{0}^{\text{enc}}$ denotes the attention mask of the background, while $\boldsymbol{m}_{k}^{\text{enc}} (1 \leq k \leq K)$ represents the attention mask of the $k$th foreground object. 

Parameters $\hat{\boldsymbol{\pi}}_{k,1:C}$ are computed based on the similarity between the canonical representations $\boldsymbol{e}_{1:C}^{\text{can}}$ and the representation of the $k$th possibly occluded object $\boldsymbol{z}_{k}^{\text{int}}$ that is predicted by a neural encoder $f_{\text{enc}}^{\text{obj}}$. In other words, $q(y_{k}|\boldsymbol{z}_{k}^{\text{ext}},\boldsymbol{x};\boldsymbol{e}_{1:C}^{\text{can}})$ can be seen as the distribution of the similarity between $\boldsymbol{e}_{1:C}^{\text{can}}$ and $\boldsymbol{e}_{k}^{\text{int}}$, and $\hat{\pi}_{c}$ will be larger if $\boldsymbol{z}_{k}^{\text{int}}$ is more similar to $\boldsymbol{e}_{c}^{\text{can}}$. The details of the computation of similarities are described in the following.

\begin{figure*}[t]
\begin{center}
\centerline{\includegraphics[width=1.5\columnwidth]{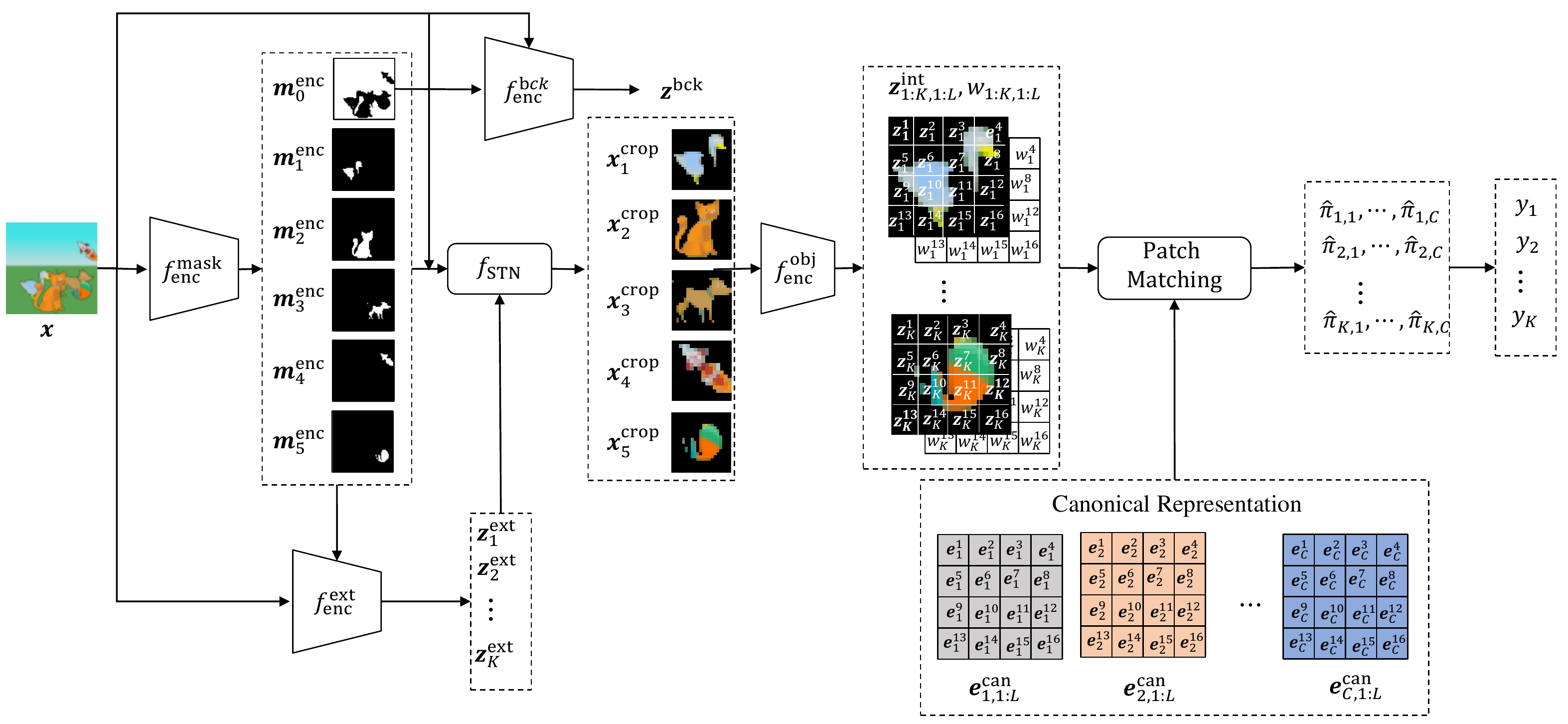}}
\caption{The procedure of identifying objects with canonical representations.  $\boldsymbol{z}^{\text{bck}}$ is sampled from the variational distribution $q(\boldsymbol{z}^{\text{bck}}|\boldsymbol{x})$ with parameters $\hat{\boldsymbol{\mu}}^{\text{bck}}$ and $\hat{\boldsymbol{\sigma}}^{\text{bck}}$. $\boldsymbol{z}_{k}^{\text{ext}}$ is sampled from the variational distribution $q(\boldsymbol{z}_{k}^{\text{ext}}|\boldsymbol{x})$ with parameters  $\hat{\boldsymbol{\mu}}_{k}^{\text{ext}}$ and $\hat{\boldsymbol{\sigma}}_{k}^{\text{ext}}$. $y_k=c$ is sampled from the categorical distribution with parameters $\hat{\boldsymbol{\pi}}_{k,1:C}$.}
\label{fig:identical}
\end{center}
\end{figure*}

\subsection{Object Identification with Canonical Representations}
Observed objects may be incomplete due to occlusion, which makes it difficult to identify objects based on the representations that are directly obtained from scene images. In order to accurately recognize objects, humans imagine the complete objects by comparing the local features of occluded objects with the local features of the memorized canonical objects that have been seen before. Inspired by this ability of humans, the proposed GOCL employs a patch-matching strategy and learns globally invariant canonical representations for canonical complete objects in order to identify the possibly occluded objects in the visual scenes. 

Aligning possibly occluded objects with canonical complete objects is challenging because the representations of complete objects and occluded objects are quite different as shown in Figure~\ref{fig:different_latent}. To address this issue, we employ a patch-matching strategy that matches the possibly occluded object to the canonical object based on local features. Specifically, each object is modeled as the composition of multiple patches, and each patch is converted to a local feature vector. The representation of a possibly occluded object or a canonical object is a set of multiple local features. The possibly occluded object can be aligned  to the corresponding canonical object by computing the similarities between the local representation of object in the visual scene and all the canonical representations that are learned from previous scenes. The procedure of identifying the object in the scene with global canonical representations is shown in Figure~\ref{fig:identical}.

The set $\boldsymbol{e}_{c}^{\text{can}}= \{\boldsymbol{e}_{c,1}^{\text{can}}, \dots,\boldsymbol{e}_{c,L}^{\text{can}} \}$ denotes the local features of the $c$th canonical object. The representation of intrinsic attributes of the $k$th object in the current scene is the set  $\boldsymbol{z}_{k}^{\text{int}}= \{\boldsymbol{z}_{k,1}^{\text{int}}, \dots,\boldsymbol{z}_{k,L}^{\text{int}}\}$. $L$ indicates the number of the patches of an object. In order to make the encoded intrinsic representation of possibly occluded object closer to the canonical representation of the corresponding canonical object, $\boldsymbol{z}_{k}^{\text{int}}$ is encouraged to contain information only about a single object, which means that only one of the $K$ objects appears on the image of the $k$th object $\boldsymbol{x}_{k}^{\text{crop}}$. However, since different objects in the scene may overlap, the cropped image $\boldsymbol{x}_{k}^{\text{crop}}$ is likely to contain multiple objects. To solve this problem, We use an attention mask $\boldsymbol{m}_{1:K}^{\text{enc}}$ which is computed using Eq.\eqref{equ:infermask} to exclude the regions of other objects. Thus, except for the $k$th object, the intrinsic representation $\boldsymbol{z}_{k}^{\text{int}}$ contains as little information as possible about other objects. In addition, some local patches of an object may be partially occluded, and it is possible that the local features of these patches are different in the intrinsic representation of this object and the corresponding canonical representation, which leads to a large difference between the intrinsic representation and the canonical representation. To solve this problem, each patch is not only converted to a vector $\boldsymbol{z}_{k,l}^{\text{int}}$ that summarizes the content in the patch, but also converted to a weight $w_{k,l}$ that measures the degree of patch occlusion. The computations of $\boldsymbol{z}_{k,1:L}^{\text{int}}$ and $\boldsymbol{w}_{k,1:L}$ are described below.
\begin{align}
    \boldsymbol{x}_{k}^{\text{crop}} &= f_{\text{STN}}(\boldsymbol{x}\odot\boldsymbol{m}_{k}^{\text{enc}}, \boldsymbol{z}_{k}^{\text{ext}}), & 1\leq k \leq K\\
    \boldsymbol{z}_{k,1:L}^{\text{int}},\boldsymbol{w}_{k,1:L}  &= f_{\text{enc}}^{\text{obj}}(\boldsymbol{x}_{k}^{\text{crop}}), &1\leq k\leq K
\end{align}
where $\odot$ denotes the element-wise multiplication.

The identification of the same object with different occlusions can be achieved by aligning the intrinsic representation $\boldsymbol{z}_{k,1:L}^{\text{int}}$ of each object observed in the visual scene image with canonical representations $\boldsymbol{e}_{1:C,1:L}^{\text{can}}$ of $C$ globally invariant canonical objects via patch matching. The patch weight $w_{k,l}$ is multiplied by the canonical representation of the patch in canonical object $\boldsymbol{e}_{c,l}^{\text{can}}$ to reduce the difference between $\boldsymbol{e}_{c,l}^{\text{can}}$ and the intrinsic representation of the patch in the possibly occluded object $\boldsymbol{z}_{k,l}^{\text{int}}$. The  similarity between intrinsic representation $\boldsymbol{z}_{k}^{\text{int}}$ and canonical representation $\boldsymbol{e}_{c}^{\text{can}}$ is measured by the similarity between each patch representation $\boldsymbol{z}_{k,l}^{\text{int}}$ and all the patch canonical representations $\boldsymbol{e}_{c,1:L}^{\text{can}}$. The mathematical expressions of patch matching is described below.
\begin{align}
    \boldsymbol{s}_{k,c,l} &= w_{k,l}\boldsymbol{e}_{c,l}^{\text{can}}\\
    \gamma_{k,c} &= (\boldsymbol{z}_{k,1:L}^{\text{int}}\boldsymbol{s}_{k,c,1:L}^{\top})/ \sqrt{D}  \\
    \hat{\gamma}_{k,c} &= \sum\nolimits_{i=1}^{L}\sum\nolimits_{j=1}^{L} \gamma_{k,c,i,j}  \\
    \hat{\pi}_{k,1}, \dots, \hat{\pi}_{k,C} &= \text{softmax}(\hat{\gamma}_{k,1}, \dots, \hat{\gamma}_{k,C})
\end{align}
In the above expressions, $D$ is the dimensionality of the intrinsic or canonical representation. $\hat{\pi}_{k,1}, \dots, \hat{\pi}_{k,C}$ are the parameters of the posterior probability distribution of categorical latent variable $y_{k}$ of the $k$th object.

\subsection{Learning of Neural Networks}
Parameters of the neural networks $f_{\text{enc}}^{\text{mask}}$, $f_{\text{enc}}^{\text{ext}}$, $f_{\text{enc}}^{\text{bck}}$, $f_{\text{enc}}^{\text{obj}}$, $g_{\text{dec}}^{\text{bck}}$, $g_{\text{dec}}^{\text{obj}}$ are learned in the variational autoencoder framework. The objective function consists of two parts and can be written as
\begin{equation}
    \label{equ:loss}
    \mathcal{L} = \mathcal{L}_{\text{ELBO}} + \alpha \mathcal{L}_{\text{mask}}
\end{equation}
In the above expression, $\mathcal{L}_{\text{ELBO}}$ is the Evidence Lower Bound (ELBO). $\mathcal{L}_{\text{mask}}$ is the regularization term of the inferred masks of objects. It is computed as the Kullback-Leibler (KL) divergence \cite{kullback1997information} between the inferred masks and decoded masks and refers to GENESIS-V2 \citep{engelcke2021genesis}. $\alpha$ is a hyper-parameter and is set to vary linearly from 0 to 3.

The computation of $\mathcal{L}_{\text{ELBO}}$ can be further expanded, and detailed expressions are described below.
\begin{align*}
    \mathcal{L}_{\text{ELBO}} =& -\mathbb{E}_{q(\boldsymbol{z}^{\text{ext}},\boldsymbol{y},\boldsymbol{z}^{\text{bck}}|\boldsymbol{x},\boldsymbol{e}_{1:C}^{\text{can}})}\big[\log p(\boldsymbol{x}|\boldsymbol{z}^{\text{ext}}, \boldsymbol{y},\boldsymbol{z}^{\text{bck}};\boldsymbol{e}_{1:C}^{\text{can}})\big] \\
    &+ D_{\text{kl}}\big(q(\boldsymbol{z}^{\text{bck}}|\boldsymbol{x})||p(\boldsymbol{z}^{\text{bck}})\big)\\
    &+ \sum\nolimits_{k=1}^{K}D_{\text{kl}}\big(q(\boldsymbol{z}_{k}^{\text{ext}}|\boldsymbol{x})||p(\boldsymbol{z}_{k}^{\text{ext}})\big)\\
    &+ \sum\nolimits_{k=1}^{K}D_{\text{kl}}(q(y_{k}|\boldsymbol{z}_{k}^{\text{ext}},\boldsymbol{x};\boldsymbol{e}_{1:C}^{\text{can}})||p(y_{k}))
\end{align*}
In $\mathcal{L}_{\text{ELBO}}$, the first term is the reconstruction loss, from the second term to the fourth term are the regularization of the variational distribution of latent variables $\boldsymbol{z}^{\text{bck}}$, $\boldsymbol{z}^{\text{ext}}$ and  $\boldsymbol{y}$, respectively. The KL divergence \cite{kullback1997information} is used to compute the above-mentioned three regularizations. $\boldsymbol{y}_k$ is a one-hot variable sampled from a categorical distribution, and the sampling operation cannot be backpropagated. In order to train the model in an end-to-end way, the discrete variable $\boldsymbol{y}_k$ is approximated using the Gumbel-softmax function  \cite{jang2016categorical} described below.
\begin{equation}
    y_{k,c} = \frac{\exp((\log \hat{\pi}_{k,c} + g_c) / \tau)}{\sum_{i=1}^{C}\exp((\log \hat{\pi}_{k,i} + g_i) / \tau)}
\end{equation}
In the above expression, $g_c \!\sim\! \text{Gumbel}(0,1)$, and $\tau$ is a temperature parameter and set to vary linearly from 30 to 1. 


\section{Experiments}
The performance of the proposed GOCL is evaluated from several aspects, including the recognition of occluded objects, the learning of canonical representations, and the decomposition of visual scenes, on four synthetic datasets. In addition, to verify the effectiveness of disentangling intrinsic and extrinsic attributes of objects and attention masks of background and objects, ablation study is conducted. Finally, we point out the limitation and clarify the future directions of the proposed GOCL.\\
\textbf{Datasets:} Four datasets are used in the experiments. These datasets are referred to as the Abstract, Animals, MNIST and CLEVR-A datasets. The scene in the Abstract dataset consists of a background and multiple objects that may occlude each other and are selected from the Absract Scene \citep{zitnick2013bringing,zitnick2013learning}. To verify that the proposed GOCL can handle scenes with complex backgrounds, five different backgrounds are used to generate scene images of Animals dataset in which objects are selected from Flying Animals \citep{yang2020learning}, which makes it harder to decompose background and objects by object-centric learning methods. In the MNIST dataset, all objects (handwritten digits selected from MNIST \citep{deng2012mnist}) are the same color, but different from the background. MNIST dataset is used to demonstrate that GOCL does not only rely on appearance to identify objects. CLEVR-A is an augmentation of CLEVR \citep{johnson2017clevr} which is a benchmark dataset for object-centric learning methods. Specifically, the numbers of object shapes in the scene are increased from the 3 to 10. CLEVR-A is used to demonstrate the feasibility of GOCL for 3D scene applications. In all datasets, the size of each image is $64 \times 64$, and the training set has $50K$ images while the test set has $1K$ images. The same objects in different scene images varies in scale, position, and occluded regions. The number of objects in the scene is $3\sim5$. The details of datasets are presented in Supplementary Materials.\\
\textbf{Baselines:} GOCL is compared against four recent state-of-the-art, two of which are representative object-centric learning methods GENESIS-V2 \cite{engelcke2021genesis} and SPACE \cite{lin2019space}, and the other two are representative prototype learning methods DTI-Sprites \cite{monnier2021unsupervised} and MarioNette \cite{smirnov2021marionette}. GENESIS-V2 is chosen because of the similar way of training. SPACE is chosen because it also disentangles extrinsic and intrinsic attributes of objects. DTI-Sprites is chosen because it can directly learning the prototype images from multi-object scene. MarioNette is chosen because it can learn a global representation of prototypes.\\
\textbf{Implementation details:} All models are trained on a single NVIDIA GeForce 3090 GPU with mini-batch size being 128. The number of canonical objects $C$ in MarioNette is set to $150$, and $C$ is $10$ in other methods. The number of slots $K$ is set to $5$ for GOCL, DTI-Sprites and SPACE, while $K$ is  set to $6$ (5 objects plus background) for GENESIS-V2 and  $2$ for MarioNette, in all experiments. All the reported results are based on $5$ evaluations on the test sets. 
\subsection{Recognition of Occluded Objects}
\textbf{Metrics:} The identification accuracy (IACC for short) of objects in the scene is used to evaluate the performance of GOCL for recognizing possibly occluded objects. As the slots are exchangeable, IACC can be calculated by using the Hungarian algorithm \citep{kuhn1955hungarian} twice: One is to match the indexes of objects in the decomposed result and the ground truth, while the other is to match the indexes of categories in the annotations and predictions.
For GENESIS-V2 and SPACE, the categories of objects are unable to be directly predicted, but calculated by the Gaussian Mixtrue Model with the object representations encoded as input.\\
\textbf{Results:} The comparison results of GOCL, GENESIS-V2, SPACE and DTI-Sprites on the three datasets are shown in Table~\ref{tab:compar}. The global representation of prototypes learned by MarioNette may come from a part of an object (shown in Figure~\ref{fig:protos}), which makes that the IACC and ARI of MarioNette are unable to be calculated. The last column of Table~\ref{tab:compar} shows that IACC of GOCL are significantly better than the compared methods on the three datasets, which suggests that GOCL have a better performance of recognizing the same objects than the compared methods. By comparing with the results of GENESIS-V2 and SPACE, it can be demonstrated that the object-centric learning model can obtain the ability to accurately identify the same objects possibly occluded in different scenes by learning globally canonical representations. It is worth noting that DTI-Sprites, has a very low IACC as the similar as GENESIS-V2 and SPACE. It may be because DTI-Sprites directly learns prototype images, and encourages to transform into all objects in the scene with a minimal number of prototype images, which can be verified from Figure~\ref{fig:protos}. Hence, it is difficult for DTI-Sprites to accurately learn prototype images for all objects. 

\begin{table}[t]
    \centering
    \small
    \caption{Performance comparison of GOCL, GENESIS-V2, SPACE and DTI-Sprites in terms of reconstruction (MSE), segmentation (ARI) and identification (IACC).}\label{tab:compar}
    \scalebox{0.82}{
    \begin{tabular}{lcccc}
      \toprule 
      \bfseries Data set & \bfseries Models &\bfseries ARI$\uparrow$ & \bfseries MSE $\downarrow$ &\bfseries IACC$\uparrow$  \\
      \midrule 
      \multirow{4}{*}{Abstract} 
    &GENESIS-V2      &0.855$\pm$3e-3    &4.0e-3$\pm$3e-5  &0.278$\pm$6e-2\\
    &SPACE           &0.626$\pm$1e-3    &1.0e-3$\pm$1e-5  &0.316$\pm$1e-3 \\
    &DTI-Sprites     &0.853$\pm$75-5    &3.1e-3$\pm$2e-5  &0.284$\pm$1e-3\\
    &GOCL            &\bfseries0.957$\pm$4e-4    &\bfseries7.9e-4$\pm$2e-5  &\bfseries0.994$\pm$6e-4\\
      \midrule
      \multirow{4}{*}{Animals} 
    &GENESIS-V2      &0.714$\pm$1e-3    &4.0e-3$\pm$6e-6   &0.308$\pm$9e-3\\
    &SPACE           &0.556$\pm$1e-3    &2.0e-3$\pm$7e-5   &0.265$\pm$2e-2\\
    &DTI-Sprites     &0.633$\pm$1e-4    &6.1e-3$\pm$8e-6   &0.279$\pm$5e-4\\
    &GOCL            &\bfseries0.934$\pm$4e-3    &\bfseries1.4e-3$\pm$2e-6   &\bfseries0.989$\pm$1e-3\\
    \midrule
      \multirow{4}{*}{MNIST} 
    &GENESIS-V2      &0.866$\pm$1e-3    &3.7e-3$\pm$2e-5   &0.299$\pm$9e-3  \\
    &SPACE           &0.863$\pm$5e-4    &2.0e-3$\pm$1e-5   &0.177$\pm$9e-3\\
    &DTI-Sprites     &0.892$\pm$1e-4    &2.9e-3$\pm$1e-5   &0.390$\pm$1e-3 \\
    &GOCL            &\bfseries0.972$\pm$6e-4    &\bfseries1.5e-3$\pm$1e-5  &\bfseries0.993$\pm$3e-4 \\
      \bottomrule 
    \end{tabular}
    }
\end{table}

\subsection{Learning of Canonical Representations}
\begin{figure}[t]
    \begin{center}
    \centerline{\includegraphics[width=0.80\columnwidth]{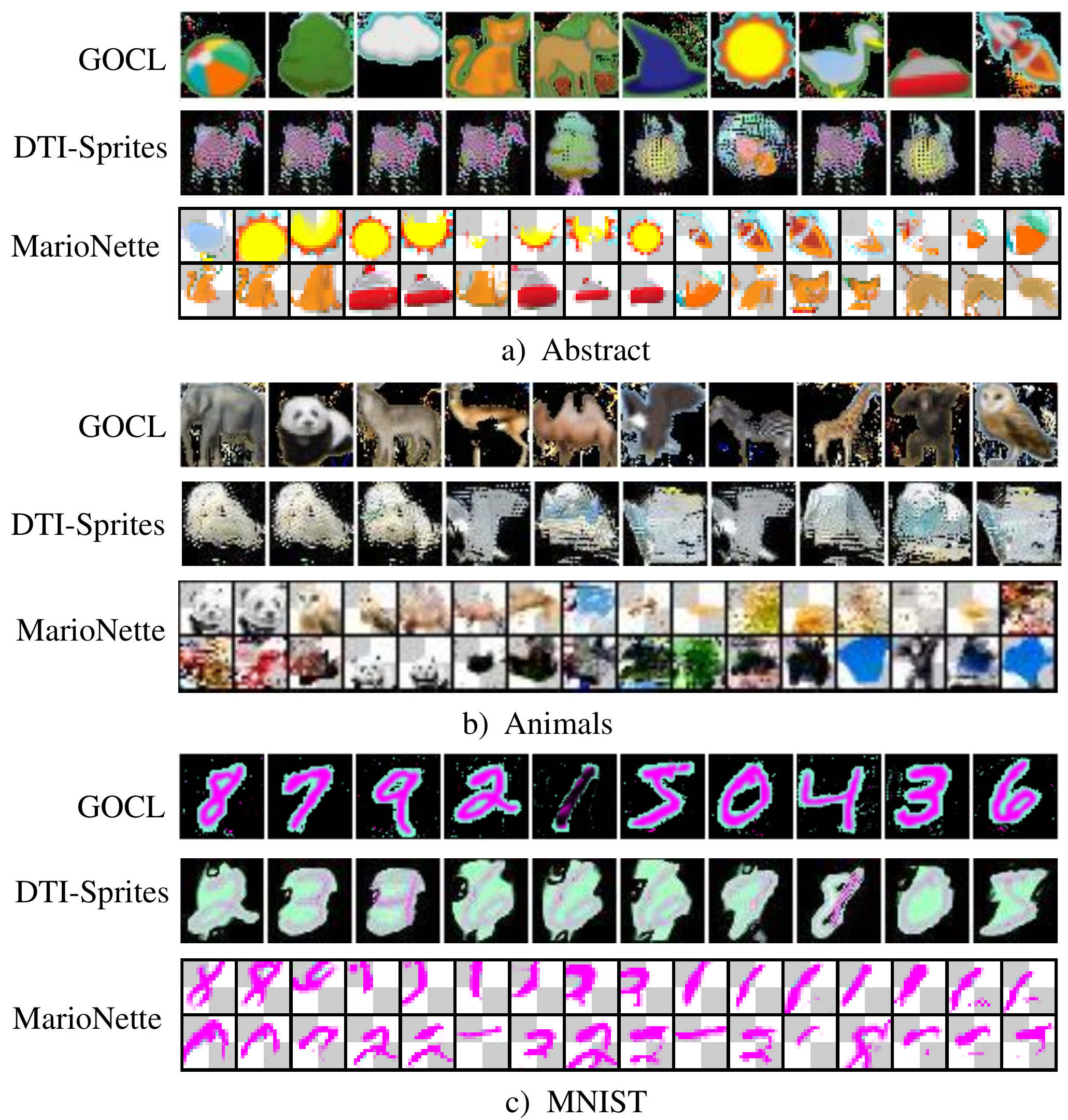}}
    \caption{The canonical object images generated by GOCL, DTI-Sprites and MarioNette on the three datasets. }
    \label{fig:protos}
    \end{center}
\end{figure}

The performance of learning globally invariant canonical representations can be reflected by the quality of canonical object images generated by a decoder with canonical representations as inputs. The generated canonical object images are shown in Figure~\ref{fig:protos}. It can be seen that the ten canonical object images in each of the Abstract, Animals and MNIST datasets are generated well, which suggests that GOCL can effectively learn the globally invariant canonical representations. It is worth noting that some images of prototypes learned by DTI-Sprites are repeated, and some of them appear to be superimposed by multiple objects. It may be because they meet the requirements of converting to all objects with a minimum number of prototypes. In addition, DTI-Sprites directly learns prototype images of objects instead of global representations like GOCL, so GOCL is more suitable for application in downstream tasks than DTI-Sprites. As can be seen from Figure~\ref{fig:protos}, the global object images (showing only a part) obtained by decoding the canonical representation learned by MarioNette are only the local of objects. It means that MarioNette can only learn the canonical representation of local regions of objects, which makes it is unable to learn the representations of canonical objects. 

\subsection{Decomposition of Visual Scenes}
\begin{figure}[t]
    \begin{center}
    \centerline{\includegraphics[width=0.8\columnwidth]{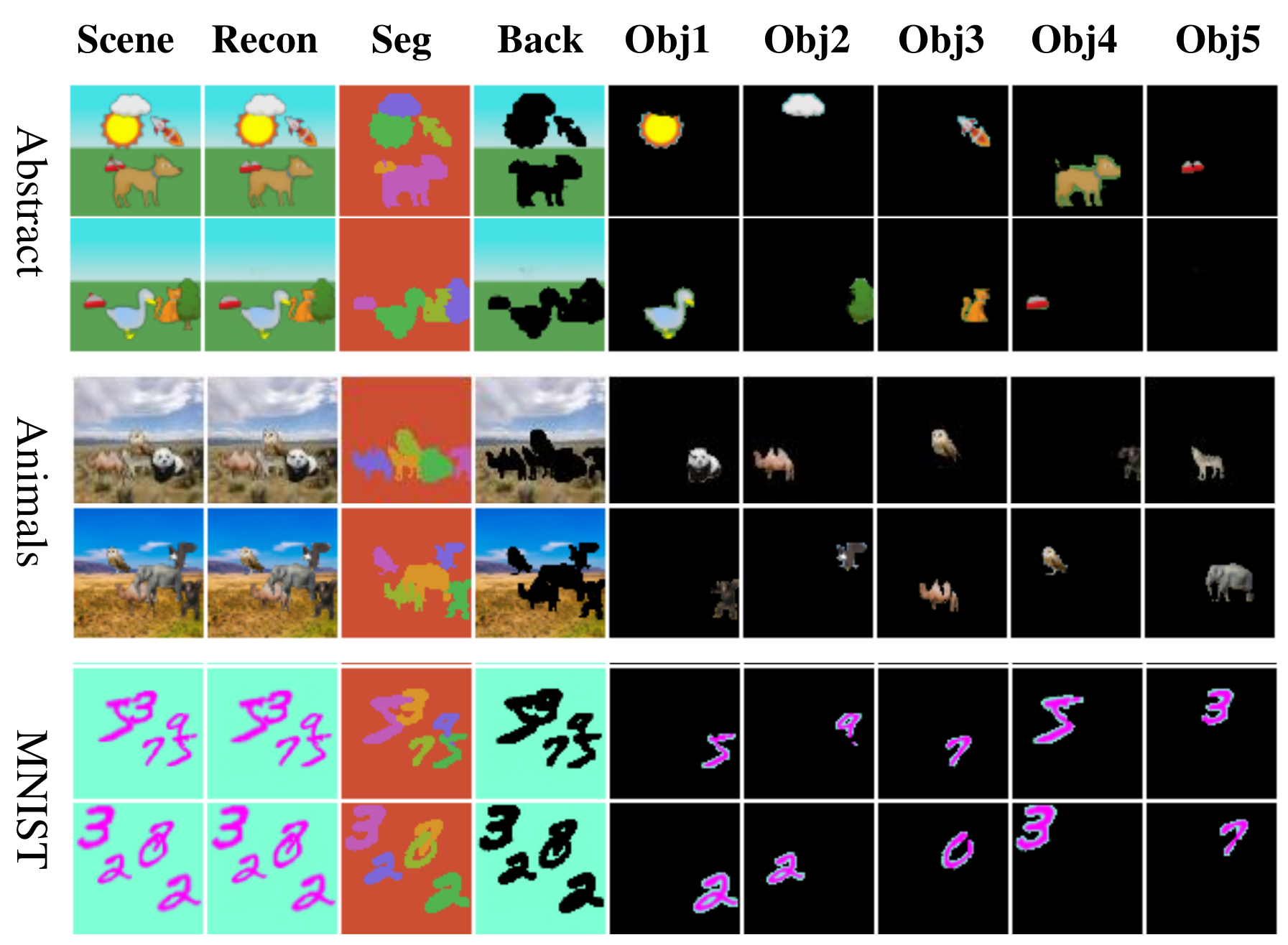}}
    \caption{The qualitative results of decomposition and reconstruction of GOCL on three datasets. For the result of each dataset, the `Scene' column  denotes input scene image, the `Recon' column represents the reconstruction of input scene image, the `Seg' column indicates the segmentation of each object in the scene , the `Back' column represents the reconstruction of background, and the columns from `Obj1' to `Obj5' represent the reconstruction of each object.}
    \label{fig:recon}
    \end{center}
\end{figure}

\textbf{Metrics:} The performance of learning object-centric representations is evaluated both quantitatively and qualitatively. The former one is usually measured in Adjusted Rand Index (ARI) \citep{pinto2007ranked}, while the latter one is reflected by the reconstructed single-object images. In our experiments, the ARI of all models on all datasets computed using ground-truth foreground pixels only. In addition, the Mean Square Error (MSE) between the input visual scene image and the reconstructed image is used to measure the reconstruction performance of the model. Models achieve better performance with higher values of ARI.\\
\textbf{Results:} The comparison of ARI and MSE of GENESIS-V2, DTI-Sprites, SPACE and GOCL on the three datasets are demonstrated in Table~\ref{tab:compar}. The third column of Table~\ref{tab:compar} shows that the ARI of GOCL is higher than other compared methods on three datasets. The fourth column of Table~\ref{tab:compar} shows that the proposed GOCL achieves the smallest MSE values on the three datasets. It means that the reconstruction performance of GOCL is significantly better than others on the three datasets. It is worth noting that the ARI of the compared methods on the Animals dataset with multiple complex backgrounds are significantly lower than those on the Abstract and MNIST datasets, while the ARI of GOCL on all three datasets are comparable. This implies that GOCL can effectively deal with the complex background in the scene. Qualitative results of scene decomposition and reconstruction of the proposed GOCL are shown in Figure~\ref{fig:recon}. It can be seen that on all three datasets, GOCL can accurately decompose objects from the scene, and reconstruct the entire scene with decomposed individual objects as well.

\subsection{Ablation Study}
In this section, we will evaluate the performance of GOCL when object attributes of intrinsic and extrinsic are not disentangled (`w/o DI') or attention masks of objects are not predicted (`w/o AM'), and verify the effectiveness of these two design choices. The proposed GOCL is denoted as `Full'. The results of three models on three datasets are shown in Table~\ref{tab:ablation}. It is demonstrated that `Full' far outperforms the other two counterparts in all the settings, which indicates that the design choices of disentangling object attributes and predicting attention masks of objects are effective and necessary. The ablation study implies that disentangling object attributes is beneficial for learning globally canonical representations for the same objects with different scales and transforms, while prediction of object and background attention masks can greatly reduce the penetration of adjacent objects into individual object representations. 

\subsection{Future Directions and Limitations}
\begin{figure}[t]
    \begin{center}
    \centerline{\includegraphics[width=0.78\columnwidth]{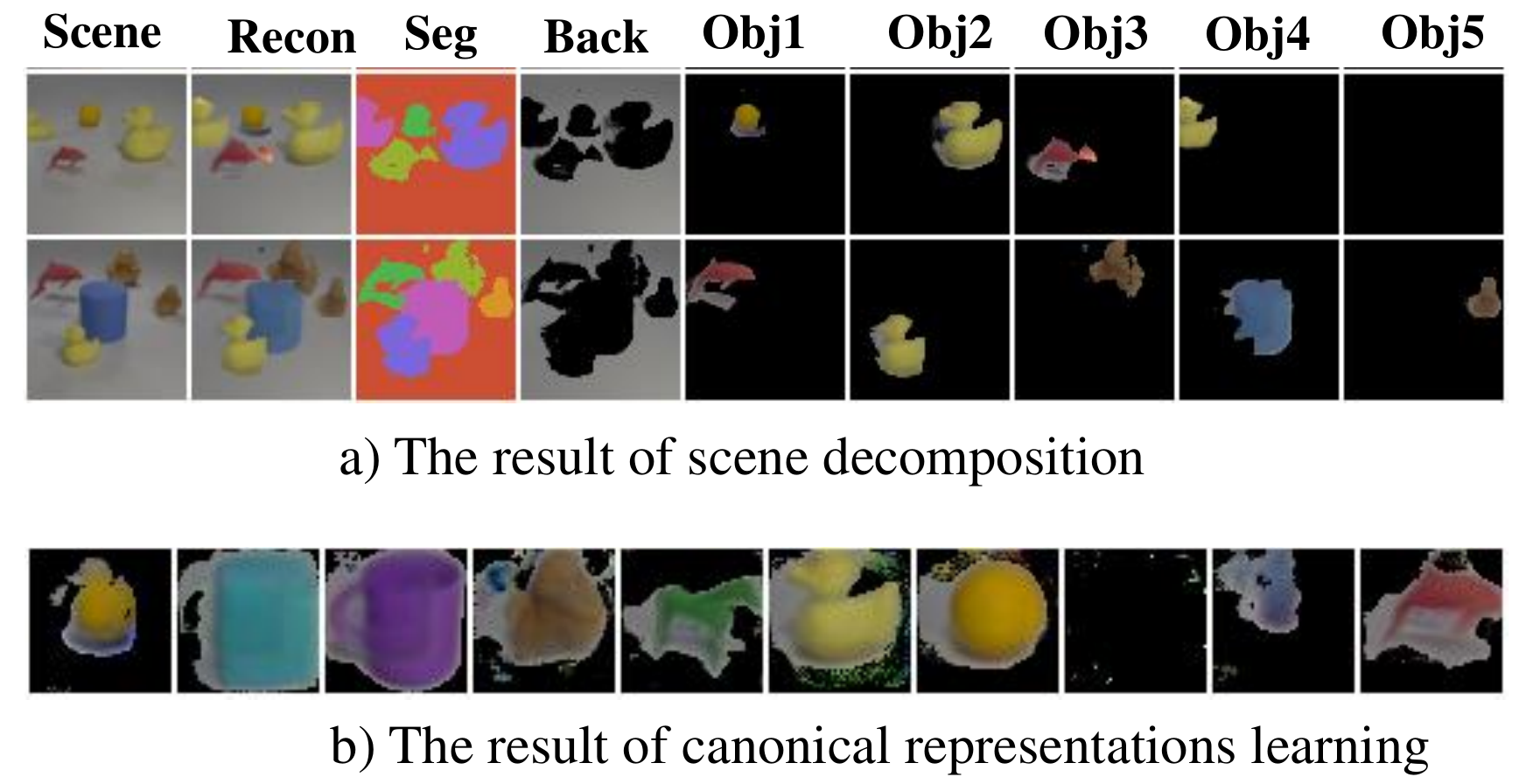}}
    \caption{The result of GOCL on CLEVR-A.}
    \label{fig:clevr-a}
    \end{center}
\end{figure}

While the GOCL is designed with 2D objects in the scene, it can be easily generalized to 3D objects in multi-view scenes. An exciting direction for future work is to model the viewpoint properties of 3D objects in order to learn a complete 3D appearance and shape representation from objects observed from different viewpoints for better application to real-world scenes. Here, we present some preliminary results. Although no modification is made to the proposed GOCL model, some interesting results is still obtained. Since the absence of the viewpoint attribute latent variable, GOCL is unable to learn the globally invariant representations of canonical objects in multi-view scenes. Hence, the recognition performance of GOCL on the same objects in 3D scenes is worse than that in 2D scenes, and the IACC of GOCL on CLEVR-A is just 0.716. Surprisingly, GOCL has a good decomposition effect on 3D scenes, and the ARI of GOCL reaches 0.921. The similar results are also verified in Figure~\ref{fig:clevr-a}. In Figure~\ref{fig:clevr-a}(a), objects in 3D scenes are well decomposed, and both individual objects and scenes are well reconstructed. In Figure~\ref{fig:clevr-a}(b), The globally invariant representations of 8 (10 in total) canonical objects are learned by GOCL. It shows that the proposed GOCL has great potential to be extended to 3D object scenes. 

\begin{table}[t]
    \small
    \centering
    \caption{The results of ablation study. }\label{tab:ablation}
    \scalebox{0.8}{
    \begin{tabular}{lcccc}
      \toprule 
      \bfseries Data set & \bfseries Model  &  \bfseries ARI$\uparrow$ &  \bfseries MSE$\downarrow$ &  \bfseries IACC$\uparrow$  \\
      \midrule 
        \multirow{4}{*}{Abstract} 
        & w/o DI  &0.654$\pm$2e-3   &4.3e-3$\pm$2e-4   &0.464$\pm$4e-3      \\
        & w/o AM  &0.247$\pm$2e-3   &7.2e-2$\pm$9e-4   &0.364$\pm$3e-3    \\
        & Full    &\bfseries0.957$\pm$4e-4    &\bfseries7.9e-4$\pm$2e-5    &\bfseries0.994$\pm$6e-4    \\
        \midrule
        \multirow{4}{*}{Animals}
        & w/o DI  &0.572$\pm$2e-3  &4.6e-3$\pm$2e-4   &0.431$\pm$4e-3    \\
        & w/o AM  &0.215$\pm$3e-3  &2.8e-2$\pm$6e-4   &0.358$\pm$3e-3    \\
        & Full    &\bfseries0.934$\pm$4e-3    &\bfseries1.4e-3$\pm$2e-6   &\bfseries0.989$\pm$1e-3\\
        \midrule
       \multirow{4}{*}{MNIST}
        & w/o DI  &0.263$\pm$2e-3   &1.5e-2$\pm$2e-3   &0.303$\pm$3e-3    \\
        & w/o AM  &0.245$\pm$2e-3   &5.7e-2$\pm$3e-4   &0.306$\pm$1e-3    \\
        & Full    &\bfseries0.972$\pm$6e-4    &\bfseries1.5e-3$\pm$1e-5  &\bfseries0.993$\pm$3e-4 \\
      \bottomrule 
    \end{tabular}
    }
\end{table}

\section{Conclusion}
In this paper, we propose an object-centric representation learning method called Global Object-Centric Learning (GOCL) to learn globally invariant canonical representations of objects without any supervision. GOCL learns better intrinsic representations by inferring the mask of each object to mitigate the interference of other objects. To be able to accurately identify the same objects that may be occluded in different scenes, a patch-matching strategy is used to align the globally invariant canonical representations with the intrinsic representations of possibly occluded objects. Experimental results demonstrate that the proposed method can identify the same object accurately and learn the compositional scene representation well. The proposed GOCL can accurately learn globally invariant canonical representations of all the objects from the data and can generate images of all the canonical objects with complete appearances and shapes. 

\bibliography{aaai23}

\newpage
\onecolumn

\section{Devrivations of ELBO}
The parameters of the proposed GOCL are learned under the framework of VAE. The posterior probability distributions $p(\boldsymbol{z}^{\text{bck}}|\boldsymbol{x})$, $p(\boldsymbol{z}^{\text{ext}}|\boldsymbol{x})$ and $p(\boldsymbol{y}|\boldsymbol{x})$ are approximated by the variational posterior probability distributions $q(\boldsymbol{z}^{\text{bck}}|\boldsymbol{x})$, $q(\boldsymbol{z}^{\text{ext}}|\boldsymbol{x})$ and $q(\boldsymbol{y}|\boldsymbol{z}^{\text{ext}},\boldsymbol{x};\boldsymbol{e}_{1:C}^{\text{can}})$. The parameters of the proposed GOCL are optimized by minimizing the KL divergence between posterior and variational posterior probability distributions of latent variables $\boldsymbol{z}^{\text{bck}}$, $\boldsymbol{z}^{\text{ext}}$ and $\boldsymbol{y}$. The expressions of this KL divergence are described in Eq.~\eqref{equ:elbo}.

\begin{align}
    \label{equ:elbo}
    D_{KL}&\big(q(\boldsymbol{z}^{\text{bck}},\boldsymbol{z}^{\text{ext}},\boldsymbol{y}|\boldsymbol{x};\boldsymbol{e}_{1:C}^{\text{can}})|| p(\boldsymbol{z}^{\text{bck}},\boldsymbol{z}^{\text{ext}},\boldsymbol{y}|\boldsymbol{x})\big) \nonumber  \\
    = &\mathbb{E}_{q(\boldsymbol{z}^{\text{bck}},\boldsymbol{z}^{\text{ext}},\boldsymbol{y}|\boldsymbol{x};\boldsymbol{e}_{1:C}^{\text{can}})}\Big[\log \frac{q(\boldsymbol{z}^{\text{bck}}\boldsymbol{z}^{\text{ext}},\boldsymbol{y}|\boldsymbol{x};\boldsymbol{e}_{1:C}^{\text{can}})}{p(\boldsymbol{z}^{\text{bck}},\boldsymbol{z}^{\text{ext}},\boldsymbol{y}|\boldsymbol{x})}\Big]\nonumber\\
    =&\mathbb{E}_{\boldsymbol{z}^{\text{bck}},q(\boldsymbol{z}^{\text{ext}},\boldsymbol{y}|\boldsymbol{x};\boldsymbol{e}_{1:C}^{\text{can}})}\big[\log q(\boldsymbol{z}^{\text{bck}},\boldsymbol{z}^{\text{ext}},\boldsymbol{y}|\boldsymbol{x};\boldsymbol{e}_{1:C}^{\text{can}})\big]  - \mathbb{E}_{q(\boldsymbol{z}^{\text{bck}},\boldsymbol{z}^{\text{ext}},\boldsymbol{y}|\boldsymbol{x};\boldsymbol{e}_{1:C}^{\text{can}})}\big[\log p(\boldsymbol{z}^{\text{bck}},\boldsymbol{z}^{\text{ext}},\boldsymbol{y}|\boldsymbol{x})\big] \\
    =&\mathbb{E}_{q(\boldsymbol{z}^{\text{bck}},\boldsymbol{z}^{\text{ext}},\boldsymbol{y}|\boldsymbol{x};\boldsymbol{e}_{1:C}^{\text{can}})}\big[\log q(\boldsymbol{z}^{\text{bck}},\boldsymbol{z}^{\text{ext}},\boldsymbol{y}|\boldsymbol{x};\boldsymbol{e}_{1:C}^{\text{can}})\big] - \mathbb{E}_{q(\boldsymbol{z}^{\text{bck}},\boldsymbol{z}^{\text{ext}},\boldsymbol{y}|\boldsymbol{x};\boldsymbol{e}_{1:C}^{\text{can}})}\big[\log p(\boldsymbol{z}^{\text{bck}},\boldsymbol{z}^{\text{ext}},\boldsymbol{y},\boldsymbol{x}) - \log p(\boldsymbol{x})\big] \nonumber\\
    =&\mathbb{E}_{q(\boldsymbol{z}^{\text{bck}},\boldsymbol{z}^{\text{ext}},\boldsymbol{y}|\boldsymbol{x};\boldsymbol{e}_{1:C}^{\text{can}})}\big[\log q(\boldsymbol{z}^{\text{bck}},\boldsymbol{z}^{\text{ext}},\boldsymbol{y}|\boldsymbol{x};\boldsymbol{e}_{1:C}^{\text{can}})\big]  - \mathbb{E}_{q(\boldsymbol{z}^{\text{bck}},\boldsymbol{z}^{\text{ext}},\boldsymbol{y}|\boldsymbol{x};\boldsymbol{e}_{1:C}^{\text{can}})}\big[\log p(\boldsymbol{z}^{\text{bck}},\boldsymbol{z}^{\text{ext}},\boldsymbol{y},\boldsymbol{x})\big]  + \log p(\boldsymbol{x}) \nonumber
\end{align}
By moving the term $\log{p(\boldsymbol{x})}$ from the right-hand side to the left-hand side, it can be shown in Eq.~\ref{equ:x-kl}.

\begin{align}
    \label{equ:x-kl}
    \log &p(\boldsymbol{x}) - D_{KL}\big(q(\boldsymbol{z}^{\text{bck}},\boldsymbol{z}^{\text{ext}},\boldsymbol{y}|\boldsymbol{x};\boldsymbol{e}_{1:C}^{\text{can}})|| p(\boldsymbol{z}^{\text{bck}},\boldsymbol{z}^{\text{ext}},\boldsymbol{y}|\boldsymbol{x})\big) \nonumber\\
    = & -\mathbb{E}_{q(\boldsymbol{z}^{\text{bck}},\boldsymbol{z}^{\text{ext}},\boldsymbol{y}|\boldsymbol{x};\boldsymbol{e}_{1:C}^{\text{can}})}\big[\log q(\boldsymbol{z}^{\text{bck}},\boldsymbol{z}^{\text{ext}},\boldsymbol{y}|\boldsymbol{x};\boldsymbol{e}_{1:C}^{\text{can}})\big]  + \mathbb{E}_{q(\boldsymbol{z}^{\text{bck}},\boldsymbol{z}^{\text{ext}},\boldsymbol{y}|\boldsymbol{x};\boldsymbol{e}_{1:C}^{\text{can}})}\big[\log p(\boldsymbol{z}^{\text{bck}},\boldsymbol{z}^{\text{ext}},\boldsymbol{y},\boldsymbol{x})\big]
\end{align}

Because $D_{KL}\big(q(\boldsymbol{z}^{\text{bck}},\boldsymbol{z}^{\text{ext}},\boldsymbol{y}|\boldsymbol{x};\boldsymbol{e}_{1:C}^{\text{can}})|| p(\boldsymbol{z}^{\text{bck}},\boldsymbol{z}^{\text{ext}},\boldsymbol{y}|\boldsymbol{x})\big)$ is greater than or equal to 0, the right-hand side of Eq.~\eqref{equ:x-kl} is a lower bound of log-likelihood $\log p(x)$ called Evidence Lower Bound (ELBO), which is denoted by $\mathcal{L}_{\text{ELBO}}^{*}$ and computed using the following expressions.

\begin{align}
    \label{equ:el}
    \mathcal{L}_{\text{ELBO}}^{*} = &-\mathbb{E}_{q(\boldsymbol{z}^{\text{bck}},\boldsymbol{z}^{\text{ext}},\boldsymbol{y}|\boldsymbol{x};\boldsymbol{e}_{1:C}^{\text{can}})}\big[\log q(\boldsymbol{z}^{\text{bck}},\boldsymbol{z}^{\text{ext}},\boldsymbol{y}|\boldsymbol{x};\boldsymbol{e}_{1:C}^{\text{can}})\big] + \mathbb{E}_{q(\boldsymbol{z}^{\text{bck}}\boldsymbol{z}^{\text{ext}},\boldsymbol{y}|\boldsymbol{x};\boldsymbol{e}_{1:C}^{\text{can}})}\big[\log p(\boldsymbol{z}^{\text{bck}},\boldsymbol{z}^{\text{ext}},\boldsymbol{y},\boldsymbol{x})\big] \nonumber\\
    =&-\mathbb{E}_{\boldsymbol{z}^{\text{bck}}, q(\boldsymbol{z}^{\text{ext}},\boldsymbol{y}|\boldsymbol{x};\boldsymbol{e}_{1:C}^{\text{can}})}\big[\log q(\boldsymbol{z}^{\text{bck}}, \boldsymbol{z}^{\text{ext}},\boldsymbol{y}|\boldsymbol{x};\boldsymbol{e}_{1:C}^{\text{can}})\big] +\mathbb{E}_{q(\boldsymbol{z}^{\text{bck}},\boldsymbol{z}^{\text{ext}},\boldsymbol{y}|\boldsymbol{x};\boldsymbol{e}_{1:C}^{\text{can}})}\big[\log p(\boldsymbol{z}^{\text{bck}}, \boldsymbol{z}^{\text{ext}},\boldsymbol{y}) + \log p(\boldsymbol{x}|\boldsymbol{z}^{\text{bck}},\boldsymbol{z}^{\text{ext}},\boldsymbol{y})\big]\nonumber\\
     =&-\mathbb{E}_{q(\boldsymbol{z}^{\text{bck}},\boldsymbol{z}^{\text{ext}},\boldsymbol{y}|\boldsymbol{x};\boldsymbol{e}_{1:C}^{\text{can}})}\big[\log \frac{q(\boldsymbol{z}^{\text{bck}},\boldsymbol{z}^{\text{ext}},\boldsymbol{y}|\boldsymbol{x};\boldsymbol{e}_{1:C}^{\text{can}})}{p(\boldsymbol{z}^{\text{bck}},\boldsymbol{z}^{\text{ext}},\boldsymbol{y})} \big] +\mathbb{E}_{q(\boldsymbol{z}^{\text{bck}},\boldsymbol{z}^{\text{ext}},\boldsymbol{y}|\boldsymbol{x};\boldsymbol{e}_{1:C}^{\text{can}})}\big[ \log p(\boldsymbol{x}|\boldsymbol{z}^{\text{bck}},\boldsymbol{z}^{\text{ext}},\boldsymbol{y})\big]\\
     =&- D_{KL}\big(q(\boldsymbol{z}^{\text{bck}},\boldsymbol{z}^{\text{ext}},\boldsymbol{y}|\boldsymbol{x};\boldsymbol{e}_{1:C}^{\text{can}})||p(\boldsymbol{z}^{\text{bck}},\boldsymbol{z}^{\text{ext}},\boldsymbol{y})\big)+\mathbb{E}_{q(\boldsymbol{z}^{\text{bck}},\boldsymbol{z}^{\text{ext}},\boldsymbol{y}|\boldsymbol{x};\boldsymbol{e}_{1:C}^{\text{can}})}\big[ \log p(\boldsymbol{x}|\boldsymbol{z}^{\text{bck}},\boldsymbol{z}^{\text{ext}},\boldsymbol{y})\big] \nonumber
\end{align}

The scene image $\boldsymbol{x}$ is assumed to consist of one background and at most $K$ independent objects. The KL divergence $D_{KL}\big(q(\boldsymbol{z}^{\text{bck}},\boldsymbol{z}^{\text{ext}},\boldsymbol{y}|\boldsymbol{x};\boldsymbol{e}_{1:C}^{\text{can}})||p(\boldsymbol{z}^{\text{bck}},\boldsymbol{z}^{\text{ext}},\boldsymbol{y})\big)$ can be written as
\begin{align}
\label{equ:kl}
    D_{KL}&\big(q(\boldsymbol{z}^{\text{bck}},\boldsymbol{z}^{\text{ext}},\boldsymbol{y}|\boldsymbol{x};\boldsymbol{e}_{1:C}^{\text{can}})||p(\boldsymbol{z}^{\text{bck}},\boldsymbol{z}^{\text{ext}},\boldsymbol{y})\big)  \nonumber\\
    =& D_{KL}\big(q(\boldsymbol{z}^{\text{bck}}|\boldsymbol{x})||p(\boldsymbol{z}^{\text{bck}})\big) 
    +\sum_{k=1}^{K}D_{KL}\big(q(\boldsymbol{z}_{k}^{\text{ext}}|\boldsymbol{x})||p(\boldsymbol{z}_{k}^{\text{ext}})\big) + \sum_{k=1}^{K}D_{KL}\big(q(y_{k}|\boldsymbol{z}_{k}^{\text{ext}},\boldsymbol{x};\boldsymbol{e}_{1:C}^{\text{can}})||p(y_{k})\big)
\end{align}
According to Eq~\eqref{equ:el} and Eq~\eqref{equ:kl}, The ELBO $\mathcal{L}_{\text{ELBO}}^{*}$ can be rewritten as
\begin{align}
    \mathcal{L}_{\text{ELBO}}^{*}=& \mathbb{E}_{q(\boldsymbol{z}^{\text{bck}},\boldsymbol{z}^{\text{ext}},\boldsymbol{y}|\boldsymbol{x},\boldsymbol{e}_{1:C}^{\text{can}})}\big[\log p(\boldsymbol{x}|\boldsymbol{z}^{\text{bck}},\boldsymbol{z}^{\text{ext}}, \boldsymbol{y};\boldsymbol{e}_{1:C}^{\text{can}})\big] \nonumber \\
    &- D_{KL}\big(q(\boldsymbol{z}^{\text{bck}}|\boldsymbol{x})||p(\boldsymbol{z}^{\text{bck}})\big) 
    - \sum_{k=1}^{K}D_{\text{kl}}\big(q(\boldsymbol{z}_{k}^{\text{ext}}|\boldsymbol{x})||p(\boldsymbol{z}_{k}^{\text{ext}})\big) 
    - \sum_{k=1}^{K}D_{\text{kl}}(q(y_{k}|\boldsymbol{z}_{k}^{\text{ext}},\boldsymbol{x};\boldsymbol{e}_{1:C}^{\text{can}})||p(y_k))
\end{align}

Since the loss function $\mathcal{L}$ needs to be minimized during training, $\mathcal{L}_{\text{ELBO}}$ in the loss function $\mathcal{L}$ is equal to the negative value of $\mathcal{L}_{\text{ELBO}}^{*}$ and is formulated as 
\begin{align}
     \mathcal{L}_{\text{ELBO}}=& -\mathbb{E}_{q(\boldsymbol{z}^{\text{bck}},\boldsymbol{z}^{\text{ext}},\boldsymbol{y}|\boldsymbol{x},\boldsymbol{e}_{1:C}^{\text{can}})}\big[\log p(\boldsymbol{x}|\boldsymbol{z}^{\text{bck}},\boldsymbol{z}^{\text{ext}}, \boldsymbol{y};\boldsymbol{e}_{1:C}^{\text{can}})\big] \nonumber\\
    &+ D_{KL}\big(q(\boldsymbol{z}^{\text{bck}}|\boldsymbol{x})||p(\boldsymbol{z}^{\text{bck}})\big) 
     + \sum_{k=1}^{K}D_{\text{kl}}\big(q(\boldsymbol{z}_{k}^{\text{ext}}|\boldsymbol{x})||p(\boldsymbol{z}_{k}^{\text{ext}})\big)
     + \sum_{k=1}^{K}D_{\text{kl}}(q(y_{k}|\boldsymbol{z}_{k}^{\text{ext}},\boldsymbol{x};\boldsymbol{e}_{1:C}^{\text{can}})||p(y_k))
\end{align}


\section{Attention Mask Prediction with $f_{\text{enc}}^{\text{mask}}$}
 The attention masks $\boldsymbol{m}_{0:K}^{\text{enc}}$ are estimated using a modified function $f_{\text{enc}}^{\text{mask}}$ based on the IC-SBP proposed in GENESIS-V2.  The IC-SBP clusters embeddings of pixels in a differentiable fashion using a stochastic stick-breaking process. The attention mask of the background cannot be distinguished from those of foreground objects in IC-SBP, on this account, IC-SBP is unable to be directly used in the proposed method. To solve this problem, the cluster center of the background is first determined in this paper. In general, the distance between two pixel embeddings belonging to the same object is smaller than the distance between two pixel embeddings belonging to different objects. Relative to all foreground objects, the background occupies the largest proportion of the entire scene area. Hence, the sum of the distances between the pixel embedding belonging the background and all the pixel embeddings in the scene is smaller than the sum of the distances between the pixel embedding belonging each foreground object and all the pixel embeddings in the scene. The clustering center of background can be determined as follows
\begin{align}
    i,j = \text{argmin}\big(\sum_{h}^{H}\sum_{w}^{W} \text{DistanceKernel}(\boldsymbol{\zeta},\zeta_{h,w})\big)
\end{align}
where $H\times W $ denotes the resolution of pixel embeddings $\zeta$ obtained by a U-Net with scene images $\boldsymbol{x}$ as input. $d_{h,w} (1\leq h \leq H, 1\leq w \leq W, HW=N)$ indicates the sum of distances between $\zeta_{h,w}$ and $\zeta_{1:H,1:W})$. The DistanceKernel is a Gaussian Kernel in this paper. The position $(i,j)~(1\leq i \leq H, 1\leq j\leq W)$ of the pixel embeddings is the cluster center of the background. When the cluster center of the background is determined, the attention masks of the background and foreground objects can be calculated in the same way as the IC-SBP.

\section{Details of Datasets}
\begin{figure*}[ht]
    \begin{center}
        \centerline{\includegraphics[width=0.6\columnwidth]{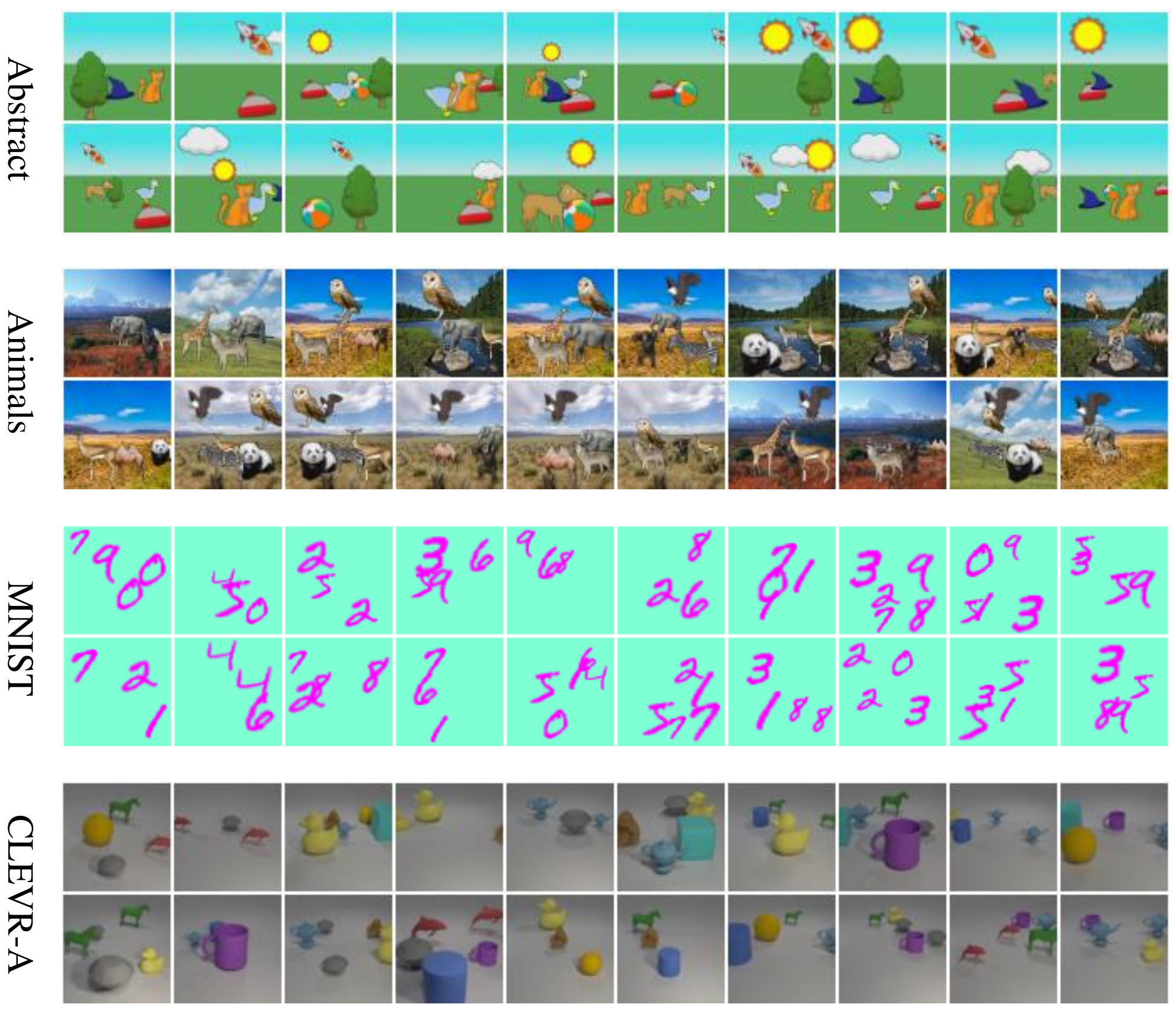}}
        \caption{The examples of multi-object scene images of four datasets used in this paper.}
        \label{fig:four-datasets}
    \end{center}
\end{figure*}

\textbf{Abstract Scene Dataset} includes 70 variants of boys and girls as well as 56 abstract objects, and is developed for the interpretation problem. In our experiments, 10 abstract objects (sun, rocket, tree, witch hat, cat, woolen hat, dog, cloud, volleyball and duck) are selected to generate scene images.\\
\textbf{Flying Animals Dataset} is composited based on 24 images of real animals and 10 images of real landscapes. In this paper, 5 images of real landscapes are  chosen as the background of the multi-object scenes, and 10 animal images (elephant, orangutan, camel, panda, owl, falcon, wolf, zebra, deer and giraffe) are selected as the foreground objects.\\
\textbf{MNIST Dataset} is a classic image recognition dataset in the field of computer vision. It consists of ten classes of handwritten digits $0 \sim 9$ with intra-class variations. In this paper, all digits regarded as foreground objects are set to the same color, and the color of background is different from the foreground object.\\
\textbf{CLEVR} is a benchmark dataset to evaluate object-centric representation learning methods, and is composed of three types of shapes (cube, cylinder and sphere) with random colors and two materials (i.e. rubber and metal). In this paper, the number of object shapes is increased from the 3 to 10 in the scene images of CLEVR-A dataset. 10 kinds of objects are cube, cylinder, sphere, diamond, dolphin, duck, horse, monkey, mug and teapot, each of which is assigned a unique color.\\
The examples of four datasets used in this paper are demonstrated in Figure~\ref{fig:four-datasets}.

\section{ Comparison of the Object Representations extracted by GOCL, GENESIS-V2 and SPACE}
\begin{figure}[ht]
    \begin{minipage}[a]{1\linewidth}
      \centering
      \centerline{\includegraphics[width=10cm]{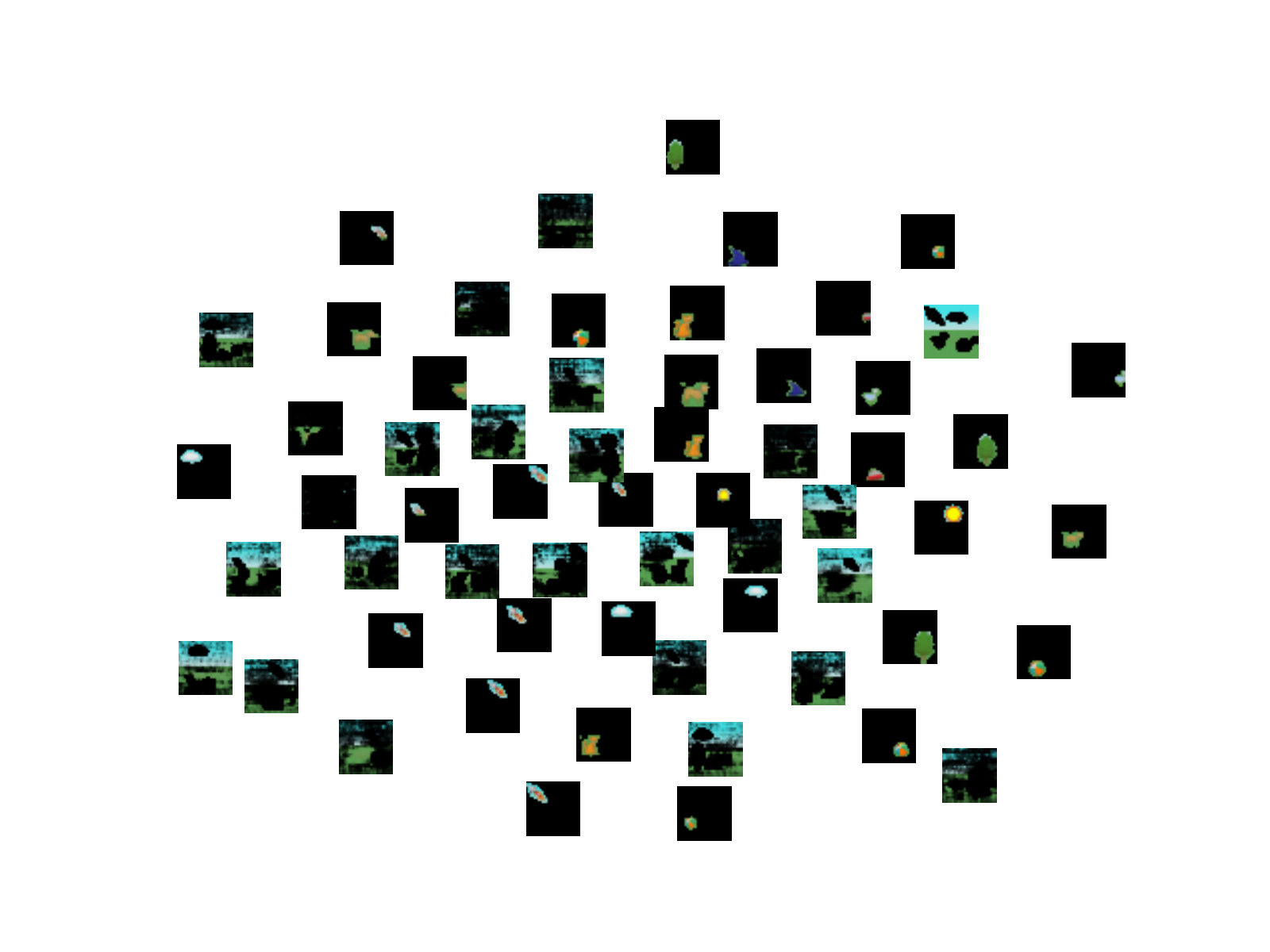}}
      \centerline{(a) GENESIS-V2}
    \end{minipage}
    \begin{minipage}[b]{1\linewidth}
      \centering
      \centerline{\includegraphics[width=10cm]{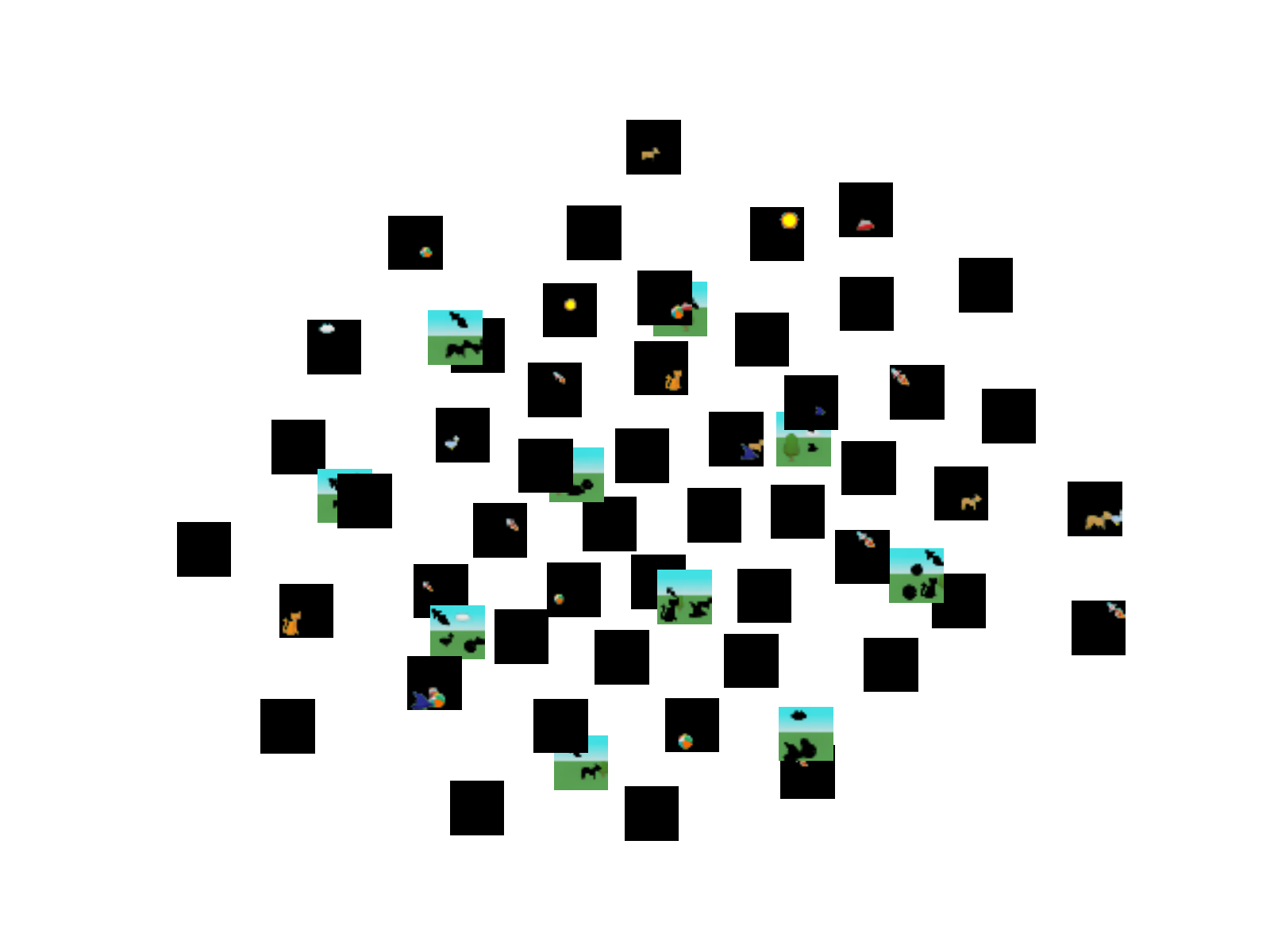}}
      \centerline{(b) SPACE}
    \end{minipage}
    \begin{minipage}[c]{1\linewidth}
        \centering
        \centerline{\includegraphics[width=10cm]{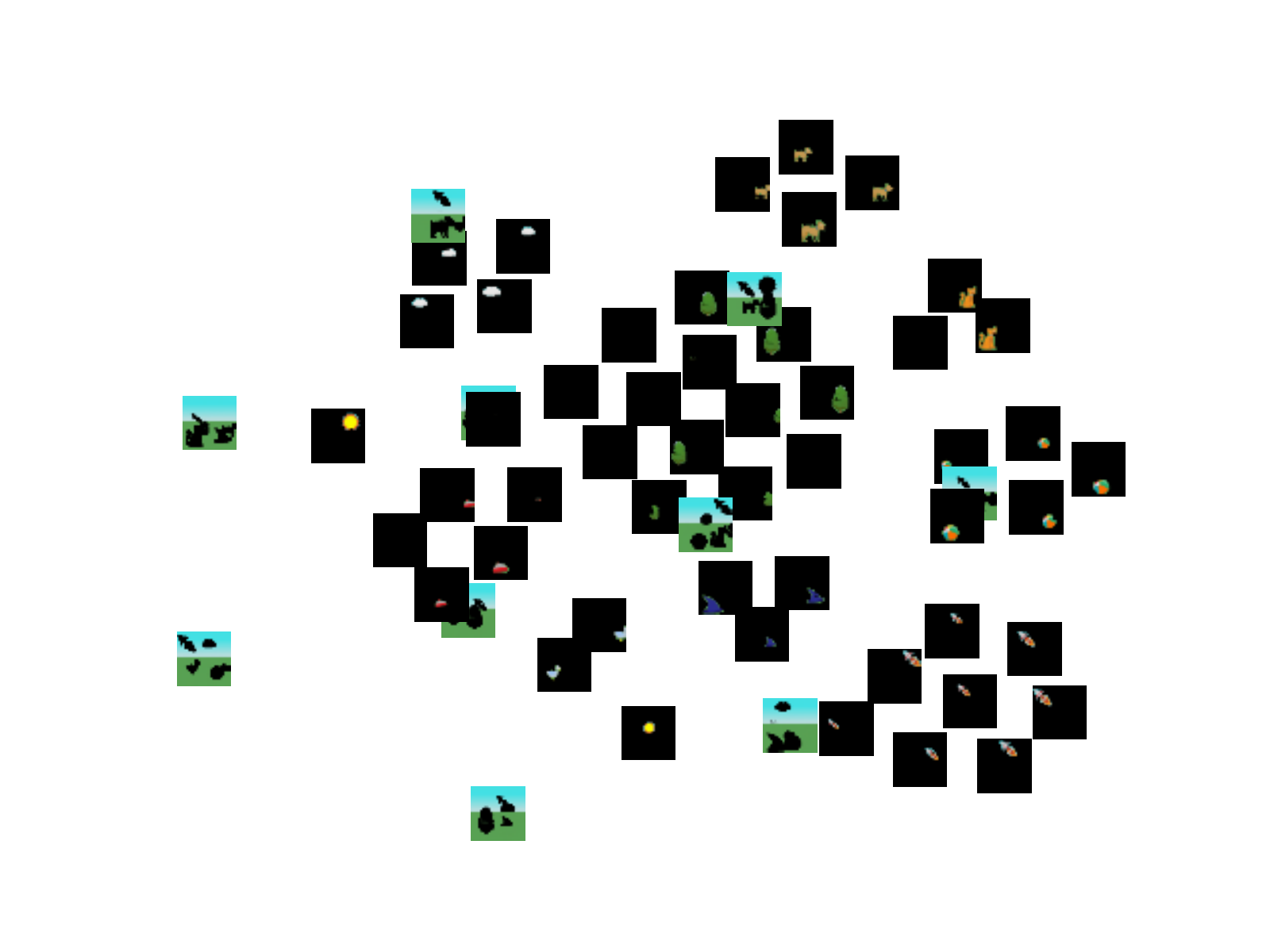}}
        \centerline{(c) GOCL}
      \end{minipage}
    \caption{Comparison of object representations of GOCL, GENESIS-V2 and SPACE on Abstract dataset.}
    \label{fig:obj-represent}
\end{figure}

For object-centric representation learning methods, the huge challenge of the recognition of the same object in multi-object scenes is that there is a large difference between the representations of the same object that may be occluded. Therefore, this paper proposes to learn a globally invariant representation for each canonical object, and identifies the same objects in different scenes by matching the representation of each object in the scene with the learned globally canonical representations via a patch-matching strategy. In the proposed GOCL, each object has a latent variable denoting category attribute, which is used to indicate the most similar globally canonical representation to it. The same objects that may be occluded in different scenes can be directed to the same globally canonical representation through inferred categorical latent variable. The distance between the representations of the same object obtained by the weighted sum of the categorical latent variables and the global canonical representations will be small. The distances in 2D space of the representations of objects in the scene extracted by GOCL GENESIS-V2 and SPACE training the model on Abstract dataset are shown in Fig~\ref{fig:obj-represent}. The object representations are embedded to a 2D space via t-SNE. The image of the object is displayed at the 2D space position corresponding to the object representation. As can be seen from Figure~\ref{fig:obj-represent}, in the models of GENESIS-V2 in Figure~\ref{fig:obj-represent}(a) and SPACE in Figure~\ref{fig:obj-represent}(b), the representations of different objects are close to each other, while the representations of the same object may be far apart. The GOCL model in Figure~\ref{fig:obj-represent}(c) showes the opposite result. It implies that the proposed GOCL model can accurately identify the same objects that may be occluded in different scenes compared to the two object-centric representation learning models, GENESIS-V2 and SPACE. In addition, the distance between the same objects is closer in the GENESIS-V2 model than that in the SPACE model. This is because the attention masks of background and foreground objects are predicted in GENESIS-V2, which minimizes the influence of adjacent objects on the extraction of each individual object representation.

\section{Comparison of Decomposition Results of GOCL and the Compared Methods}
\begin{figure}[ht]
    \begin{center}
        \centerline{\includegraphics[width=0.6\columnwidth]{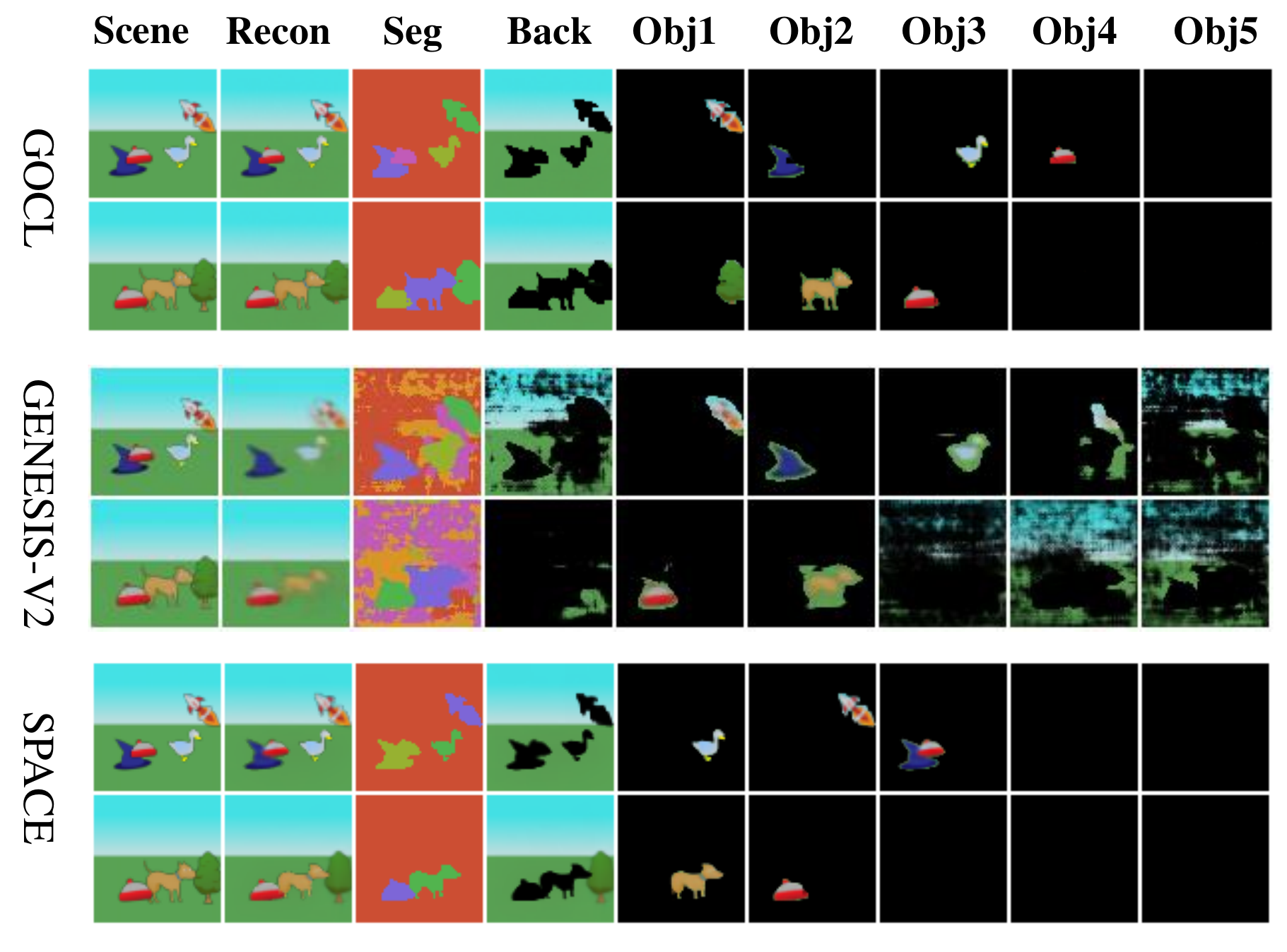}}
        \caption{Comparison of decomposition results of GOCL, GENESIS-V2 and SPACE on Abstract dataset.}
        \label{fig:test_recon_abstract}
    \end{center}
\end{figure}

\begin{figure}[ht]
    \begin{center}
        \centerline{\includegraphics[width=0.6\columnwidth]{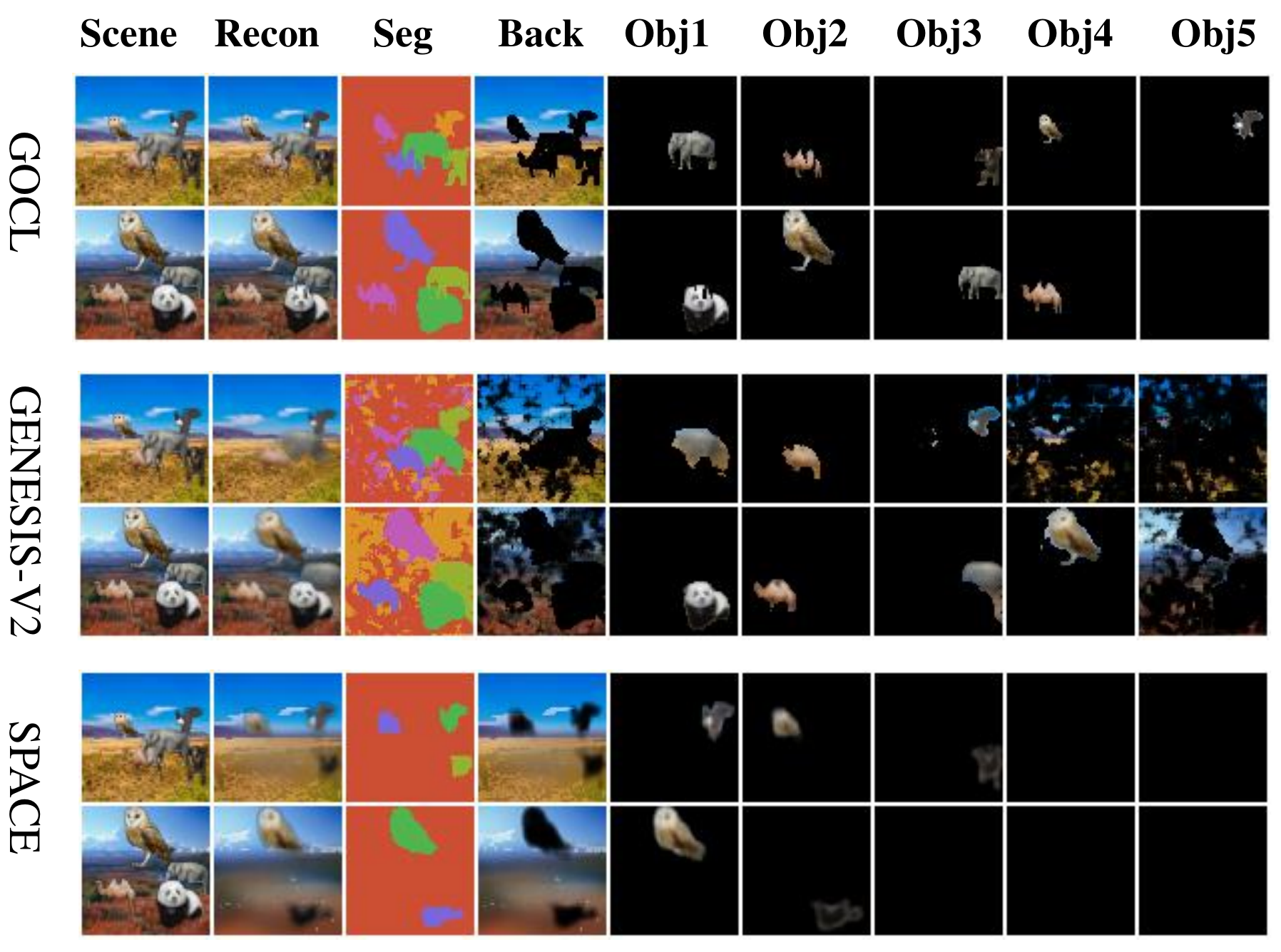}}
        \caption{Comparison of decomposition results of GOCL, GENESIS-V2 and SPACE on Animals dataset.}
        \label{fig:test_recon_animals}
    \end{center}
\end{figure}

\begin{figure}[ht]
    \begin{center}
        \centerline{\includegraphics[width=0.6\columnwidth]{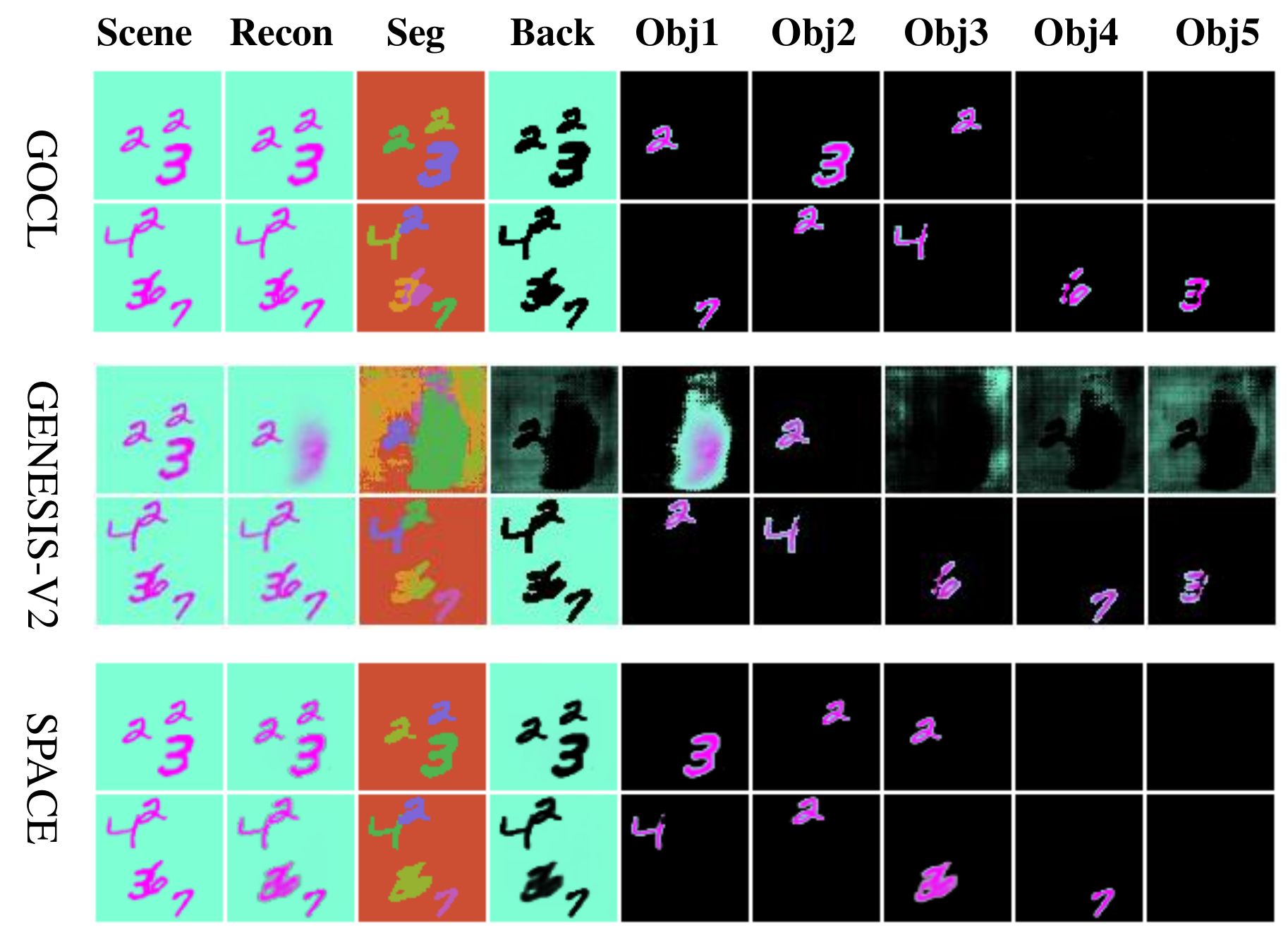}}
        \caption{Comparison of decomposition results of GOCL, GENESIS-V2 and SPACE on MNIST dataset.}
        \label{fig:test_recon_mnist}
    \end{center}
\end{figure}

\textbf{Scene Decomposition}\quad GENESIS-V2 and SPACE, as two object-centric representation leraning methods, are compared with the proposed GOCL in terms of scene decomposition on Abstract, Animals and MNIST datasets, which of the results are shown in Figure~\ref{fig:test_recon_abstract},~\ref{fig:test_recon_animals} and ~\ref{fig:test_recon_mnist}, respectively. As can be seen from Figure~\ref{fig:test_recon_abstract},~\ref{fig:test_recon_animals} and ~\ref{fig:test_recon_mnist}, the decomposition performances of GOCL on three datasets are better than the compared methods. As can be seen from Figure~\ref{fig:test_recon_animals}, it is difficult to separate foreground objects from the complex background and reconstructe the scene for GENESIS-V2 and SPACE, while it is hardly difficult for the proposed GOCL. In addition, the representations of foreground objects and background can not be distinguished in GENESIS-V2, while GOCL can guarantee that the representation of background is extracted in the first slot. Therefore, the images of 'Back' column in Figure ~\ref{fig:test_recon_abstract},~\ref{fig:test_recon_animals} and ~\ref{fig:test_recon_mnist} reconstructed by GOCL are always the backgrounds, while the images of 'Back' column in Figure ~\ref{fig:test_recon_abstract},~\ref{fig:test_recon_animals} and ~\ref{fig:test_recon_mnist} reconstructed by GENESIS-V2 may be foreground objects.

\begin{figure}[h]
    \begin{center}
        \centerline{\includegraphics[width=0.8\columnwidth]{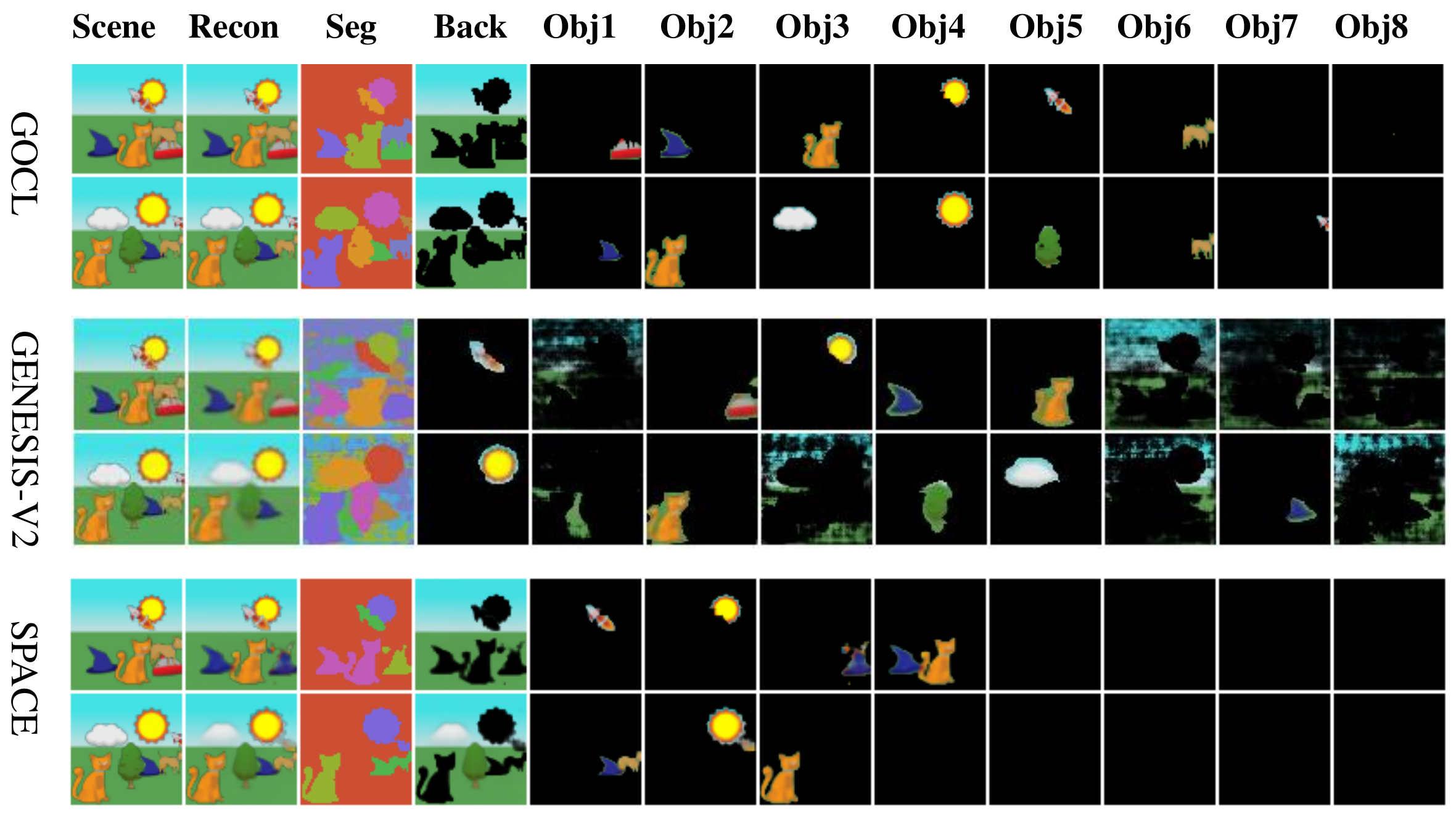}}
        \caption{Comparison of decomposition results of GOCL, GENESIS-V2 and SPACE on Abstract dataset.}
        \label{fig:general_recon_abstract}
    \end{center}
\end{figure}

\begin{figure}[h]
    \begin{center}
        \centerline{\includegraphics[width=0.8\columnwidth]{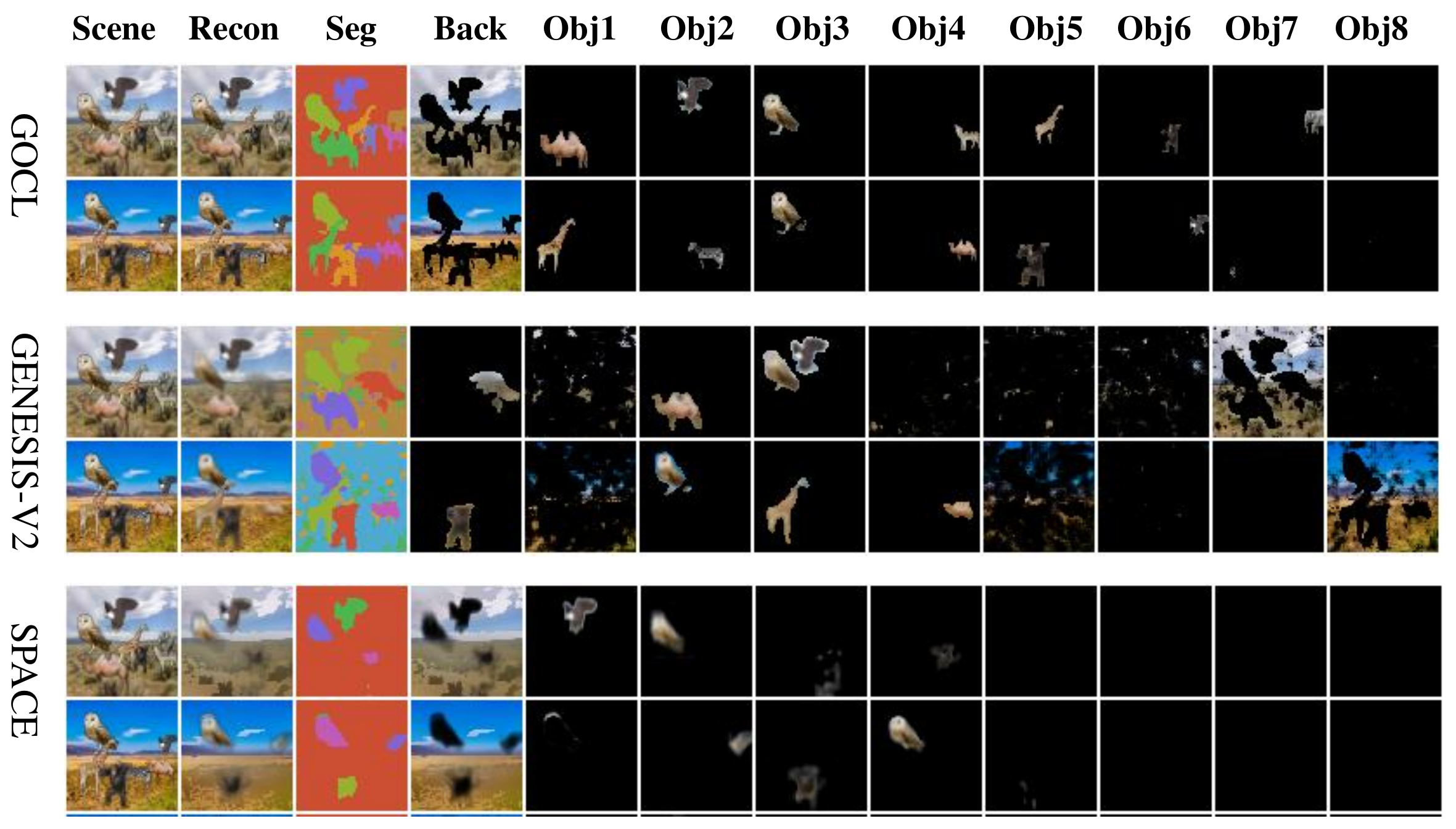}}
        \caption{Comparison of decomposition results of GOCL, GENESIS-V2 and SPACE on Animals dataset.}
        \label{fig:general_recon_animals}
    \end{center}
\end{figure}

\begin{figure}[h]
    \begin{center}
        \centerline{\includegraphics[width=0.8\columnwidth]{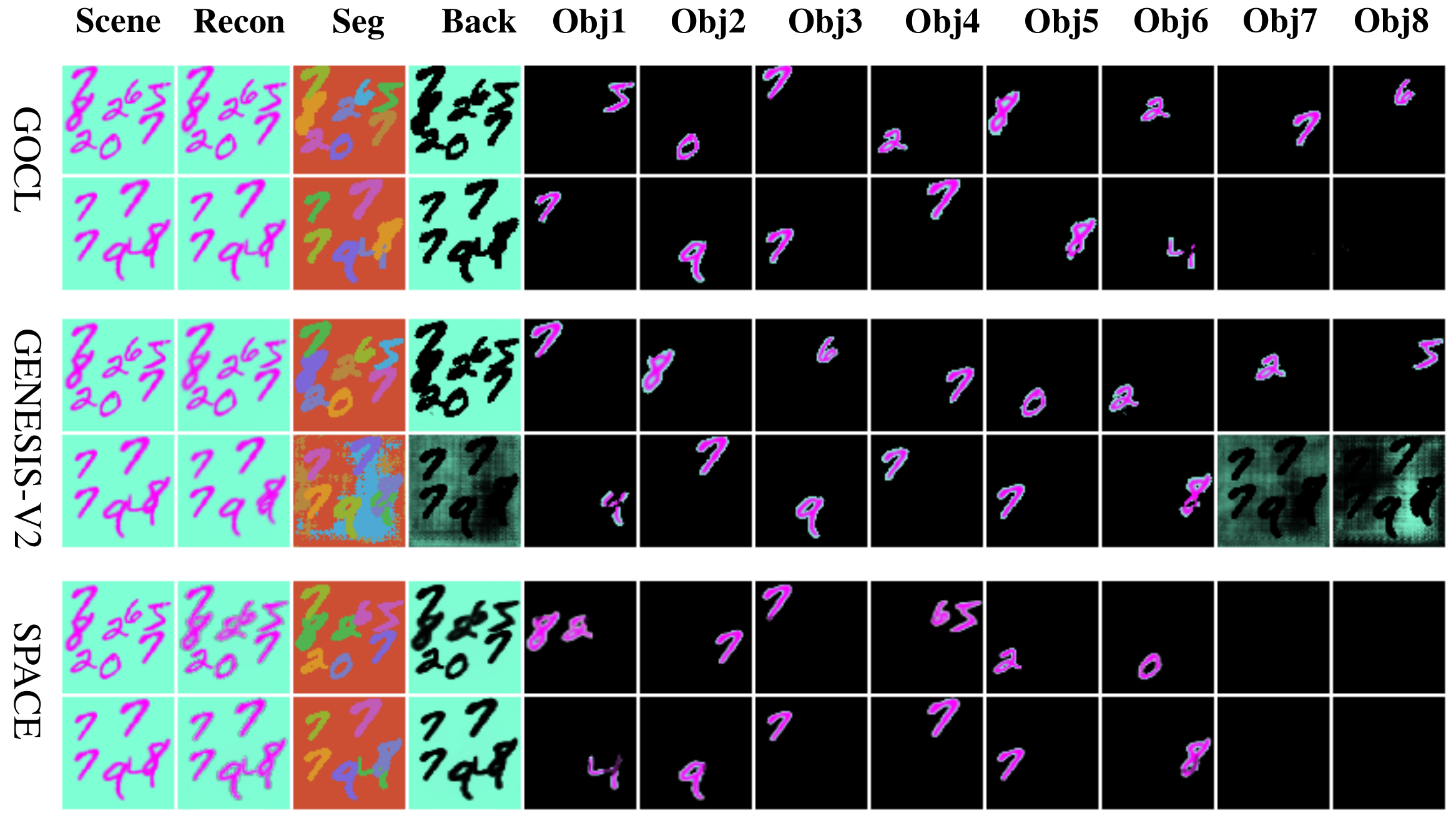}}
        \caption{Comparison of decomposition results of GOCL, GENESIS-V2 and SPACE on MNIST dataset.}
        \label{fig:general_recon_mnist}
    \end{center}
\end{figure}

\begin{figure}[h]
    \begin{center}
        \centerline{\includegraphics[width=\columnwidth]{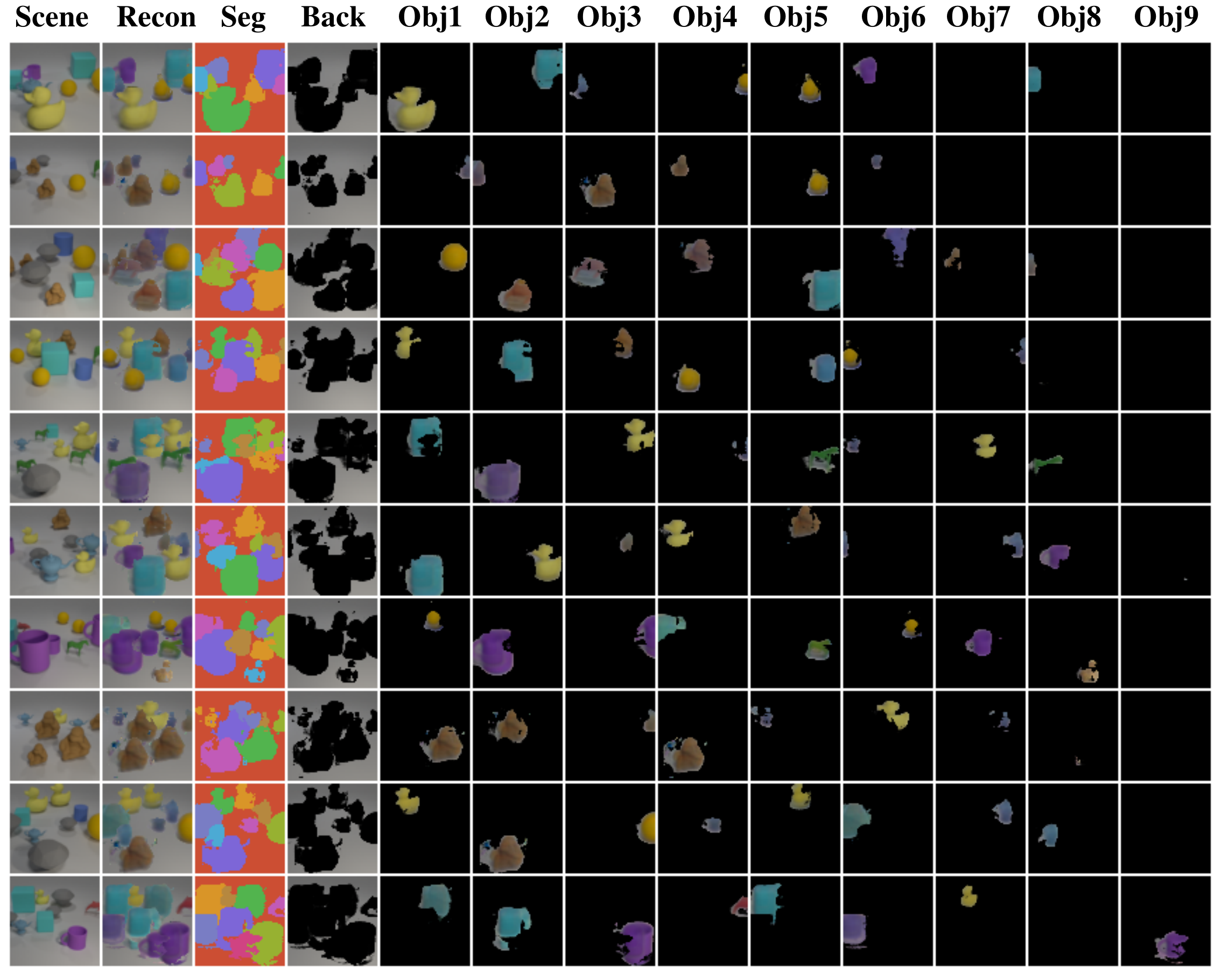}}
        \caption{Decomposition results of GOCL on CLEVR-A dataset.}
        \label{fig:general_recon_clevr-a}
    \end{center}
\end{figure}

\textbf{Generalization of GOCL} \quad In this part, the qualitative and quantitative results are used to analyse the generalization ability of the proposed GOCL. The qualitative results of scene decompositions of GOCL, GENESIS-V2 and SPACE on Abstract, Animals and MNIST datasets are shown in Figure~\ref{fig:general_recon_abstract},~\ref{fig:general_recon_animals} and ~\ref{fig:general_recon_mnist}, respectively. The quantitative results of ARI, MSE and IACC of GOCL and the compared methods on Abstract, Animals and MNIST datasets are shown in Table~\ref{tab:general-2D}. For Abstract, Animals and MNIST dataset, the models of GOCL and SPACE are trained with $K=5$ and tested with $K=8$, while the GENESIS-V2 model is trained with $K=6$ and tested with $K=9$. It can be found from Figure~\ref{fig:general_recon_abstract},~\ref{fig:general_recon_animals} and ~\ref{fig:general_recon_mnist} that as soon as the number of objects in the scene increases, GOCL can still accurately decompose all foreground objects and backgrounds in the scene, which is difficult for GENESIS-V2 and SPACE. Specifically, GENESIS-V2 tends to decompose the background into multiple slots, while SPACE easily decompose multiple objects into one slot. The results in the Table~\ref{tab:general-2D} show that GOCL outperforms the comparison methods on all metrics, especially in IACC. The above results show that GOCL has good generalization on Abstract, Animals and MNIST datasets. The qualitative and quantitative results of GOCL trained with $K=5$ and tested with $K=9$ on CLEVR-A dataset are shown in Figure~\ref{fig:general_recon_clevr-a} and Table\ref{tab:general-3D}. As can be seen from the Figure~\ref{fig:general_recon_clevr-a}, the proposed GOCL can accurately decompose foreground objects and background from the scene and reconstruct the scene well to a certain extent. The results in Figure~\ref{fig:general_recon_clevr-a} and Table\ref{tab:general-3D} show that GOCL also has good generalization ability to 3D scene dataset, which shows that it has the potential to be applied to 3D scenes after modeling the viewpoint attribute of objects in the 3D scene.

\begin{table}[ht]
    \centering
    \caption{Comparison of quantitative results. }
    \label{tab:general-2D}
    \begin{tabular}{lcccc}
      \toprule 
      \bfseries Data set & \bfseries Model & \bfseries ARI$\uparrow$  & \bfseries MSE$\downarrow$ &  \bfseries IACC$\uparrow$  \\
      \midrule 
        \multirow{4}{*}{Abstract} 
        &GENESIS-V2    &0.852$\pm$9e-4  &7.5e-3$\pm$2e-5  &0.251$\pm$9e-2     \\
        &SPACE         &0.538$\pm$3e-4  &3.0e-3$\pm$2e-5  &0.223$\pm$2e-3     \\
        &GOCL          &\bfseries0.923$\pm$8e-4      &\bfseries2.1e-3$\pm$2e-5      &\bfseries0.962$\pm$1e-3     \\
        \midrule
        \multirow{4}{*}{Animals}
        &GENESIS-V2    &0.629$\pm$1e-3  &6.8e-3$\pm$1e-5  &0.209$\pm$6e-2     \\
        &SPACE         &0.483$\pm$5e-4  &1.2e-2$\pm$2e-5  &0.207$\pm$5e-3     \\
        &GOCL          &\bfseries0.927$\pm$2e-4  &\bfseries2.0e-3$\pm$1e-5  &\bfseries0.981$\pm$4e-4     \\
        \midrule
      \multirow{4}{*}{MNIST}
        &GENESIS-V2    &0.922$\pm$1e-3  &6.8e-3$\pm$8e-5  &0.194$\pm$1e-2      \\ 
        &SPACE         &0.831$\pm$4e-4  &5.0e-3$\pm$1e-5  &0.206$\pm$1e-2     \\
        &GOCL          &\bfseries0.958$\pm$3e-4  &\bfseries2.6e-3$\pm$3e-5  &\bfseries0.985$\pm$1e-3     \\
      \bottomrule 
    \end{tabular}
\end{table}

\begin{table}[ht]
    \centering
    \caption{Generalization results of GOCL on CLEVR-A dataset. GOCL is trained with $K=5$, and tested with $K=9$. }
    \label{tab:general-3D}
    \begin{tabular}{lccc}
      \toprule 
      \bfseries Model & \bfseries ARI$\uparrow$  & \bfseries MSE$\downarrow$ &  \bfseries IACC$\uparrow$  \\
      \midrule 
       GOCL          &0.863$\pm$8e-4      &5.3e-3$\pm$4e-5      &0.661$\pm$2e-3     \\
      \bottomrule 
    \end{tabular}
\end{table}

\section{Choices of Hyperparameters }
\textbf{DTI-Sprites}\quad DTI-Sprites is trained with the default hyperparameters of CLEVR6 described in the original paper  for all datasets except: 1) the number of sprites is 11 (including 10 objects and one empty sprite); 2) the maximum number of objects in the scene is 5; 3) the number of training data is 50K; 4) the size of training image is $ 64\times 64$. \\
\textbf{MarioNette}\quad MarioNette is trained with the default hyperparameters described in the original paper for all datasets
except: 1) the number of training data is 50K; 2) the number of training data is 50K; 3) the size of training image is $ 64\times 64$. \\
\textbf{SPACE}\quad SPACE is trained with the default hyperparameters described in the original paper
except: 1) the number of training data is 50K and the batch size is 16; 2) the grid of each image is $8\times 8$ for all datasets; 3) the componnets of background $K$ is 1 for Abstract and MNIST datasets and $K$ is 2 for Animals dataset with complex backgroundands; 4) the dimensionality of the background latent variable is 4 for Abstract and Animals datasets and 1 for MNIST dataset ; 5) the background modeling is much simpler than foreground modeling, that is, the hidden channels of the network are much smaller. We chose 32 as the hidden channels of the background network. \\
\textbf{GENESIS-V2}\quad GENESIS-V2 is trained with the default hyperparameters described in the original paper
except: 1) the number of training data is 50K and the batch size is 128; 2) image size is $64 \times 64$ for all datasets; 3) the number of slots $K$ is 6 (including 5 objects and one background) ; 4) the learning rate is set to 2e-4 for learning from the Abstract and Mnist datasets, and 1e-3 for learning from the Animals dataset; 5) we slightly increase the coefficients of GECO objective used by GENESIS-V2, in order to optimize the KL divergence in a shorter time. The GECO goal (coefficient) is adjusted from 0.5655 to 0.566.\\
\textbf{GOCL}\quad In the generative model, the parameters of categorical distribution are $\pi_1 =\cdots = \pi_C=1/10 ~(C=10)$ for all datasets. The parameters of prior distribution of latent variables describing extrinsic attributes $\boldsymbol{\mu}^{\text{ext}}$ and $\boldsymbol{\sigma}^{\text{ext}}$ are chosen to be $[-1.0,-1.0,0.0,0.0]$ and $[0.2,0.2,0.5,0.5]$ for Abstract and Animals datasets, and to be $[-0.4,-0.4,0.0,0.0]$ and $[0.2,0.2,0.5,0.5]$ for MNIST dataset. The coefficient of $\mathcal{L}_{\text{mask}}$ $\alpha$ is set to vary linearly from 0 to 3 for Abstract and MNIST datasets, and from 0 to 0.01 for Animals dataset. The coefficient of KL divergence between the variational distribution and prior distribution of the latent variable of background is set to 0.1 for Animals dataset and is set to 1 for Abstract and MNIST datasets. The standard deviation of the likelihood function is $\sigma=0.3$. The object decoder $g_{\text{dec}}^{\text{obj}}$ is trained with the hyperparameters described in the official code repository \footnote{https://github.com/applied-ai-lab/genesis} of GENESIS-V2 except that the output size is $4\times 48\times 48$. The Spatial Transform Network (STN) $f_{\text{STN}}^{-1}$ is referred from the official code repository \footnote{https://github.com/jinyangyuan/infinite-occluded-objects} of GMIOO and used to convert a $48\times 48$ canonical object image into a $64\times 64$ object layer that is used to compose the scene image. In the inference model, $D=64$ is the dimension of global canonical representations of canonical object images and intrinsic representations of possible occluded objects in the current scene image. The dimensionality of latent variable of background is 4. The IC-SBP module is trained with the hyperparameters referred from the official code repository \footnote{https://github.com/applied-ai-lab/genesis} of GENESIS-V2. The extrinsic encoder $f_{\text{enc}}^{\text{ext}}$ 
used to predict the parameters of variational posterior probability distribution of the latent variable indicating extrinsic attributes is implemented using four convolutional neural networks (CNN) followed by three linear layer with output dimension of 8 (the extrinsic attributes latent variable dimension is 4). The specific network structure of the extrinsic encoder $f_{\text{enc}}^{\text{ext}}$ is shown in Table~\ref{tab:enc-ext}. $f_{\text{STN}}$ is an inverse process of $f_{\text{STN}}^{-1}$, and used to crop an object image from the multi-object scene image and convert it to a canonical image. The input image size of $f_{\text{STN}}$ is $64\times 64$, and the output image size is $48\times 48$. The object encoder $f_{\text{enc}}^{\text{obj}}$ consists of one layer normalization (LayerNorm) and two linear layers and its details are shown in Table~\ref{tab:enc-obj}. Each canonical object can be divided into $L~(L=16)$ patches each of which can be converted into the intrinsic representation $\boldsymbol{z}_{k,l}^{\text{int}} (1\leq l \leq L)$ with 64 dimensions and the weight $w_{k,l} (1\leq l \leq L)$ with 1 dimension. The background encoder $f_{\text{enc}}^{\text{bck}}$ consists of one activation function ReLU and two linear layers and its details are shown in Table~\ref{tab:enc-bck}. The background decoder $f_{\text{dec}}^{\text{bck}}$ consists of one activation function ReLU and two linear layers and its details are shown in Table~\ref{tab:dec-bck}.

\begin{table}[ht]
    \centering
    \caption{The details of the extrinsic encoder. Conv $1\times 1$ denotes the CNN layer with kernel size of $1 \times 1$. MLP indicate the Linear layer.}\label{tab:enc-ext}
    \begin{tabular}{cccc}
      \toprule 
      \bfseries Type & \bfseries Size/Channels  & \bfseries Activation &  \bfseries Comment \\
      \midrule 
      Conv $1\times 1$  & 128  & ReLU  & 1  \\
      Conv $3\times 3$  & 128  & ReLU  & 2  \\
      Conv $3\times 3$  & 64   & ReLU  & 2  \\
      Conv $1\times 1$  & 64   & ReLU  & 1  \\
      MLP               & 512  & ReLU  & -  \\
      MLP               & 256  & ReLU  & -  \\
      MLP               & 8    & -     & -  \\
      
      \bottomrule 
    \end{tabular}
\end{table}

\begin{table}[ht]
    \centering
    \caption{The details of the object encoder. The canonical image size is $48\times 48$.  MLP represents the Linear layer.}\label{tab:enc-obj}
    \begin{tabular}{ccc}
      \toprule 
      \bfseries Type & \bfseries Size/Channels  & \bfseries Activation \\
      \midrule 
      LayerNorm     & 3*48*48  & -     \\
      MLP           & 2*64     & ReLU   \\
      MLP           & 16*(64+1)     & -      \\
      
      \bottomrule 
    \end{tabular}
\end{table}

\begin{table}[ht]
    \centering
    \caption{The details of the background encoder. The image size is $64\times 64$.  MLP represents the Linear layer.}\label{tab:enc-bck}
    \begin{tabular}{ccc}
      \toprule 
      \bfseries Type & \bfseries Size/Channels  & \bfseries Activation \\
      \midrule 
      MLP           & 256     & ReLU   \\
      MLP           & 8     & -      \\
      \bottomrule 
    \end{tabular}
\end{table}

\begin{table}[ht]
    \centering
    \caption{The details of the background decoder. The image size is $64\times 64$.  MLP represents the Linear layer.}\label{tab:dec-bck}
    \begin{tabular}{ccc}
      \toprule 
      \bfseries Type & \bfseries Size/Channels  & \bfseries Activation \\
      \midrule 
      MLP           & 256     & ReLU   \\
      MLP           & 4*64*64     & -      \\
      \bottomrule 
    \end{tabular}
\end{table}

\end{document}